\documentclass[11pt]{article}
\PassOptionsToPackage{table}{xcolor}
\usepackage[preprint]{acl}
\usepackage{times}
\usepackage{latexsym}
\usepackage[T1]{fontenc}
\usepackage[utf8]{inputenc}
\usepackage{microtype}
\usepackage{inconsolata}
\usepackage{graphicx}
\usepackage{booktabs}
\usepackage{amsmath}
\usepackage{amssymb}
\usepackage{xcolor}
\usepackage{tabularx}
\usepackage{enumitem}
\usepackage{tikz}
\usetikzlibrary{arrows.meta,positioning,fit,calc,shapes.misc,decorations.pathreplacing}
\usepackage{caption}
\usepackage{xurl}
\graphicspath{{figs/}}
\newcommand{\dmtwo}{\mbox{GLiDE~V1.2}}
\newcommand{\dmone}{\mbox{GLiDE~V1.1}}
\newcommand{\basemodel}{Qwen3.8-27B}

\newcommand{\DI}{Decision Index}

\definecolor{dmdark}{HTML}{9A3412}
\definecolor{dmorange}{HTML}{E8590C}
\definecolor{dmgray}{HTML}{78716C}
\definecolor{dmlight}{HTML}{FDEBDF}

\newif\ifshowchanges \showchangesfalse
\ifshowchanges
  \DeclareRobustCommand{\new}[1]{{\color{blue}#1}}
\else
  \DeclareRobustCommand{\new}[1]{#1}
\fi
\definecolor{pipeblue}{HTML}{E8F0FA}\definecolor{pipenavy}{HTML}{1D3A68}
\definecolor{pipeorange}{HTML}{FDE1BF}\definecolor{pipebrown}{HTML}{7D3A01}

\title{GLiDE: A Fast, Calibrated Decision Model Distilled from \\ Reasoning Teachers}

\author{Sahibzada Allahyar \quad
  Julia White \quad
  Dhruv Atreja \quad \\
  \textbf{George Hurn-Maloney} \quad
  \textbf{Urchade Zaratiana} \\
  Fastino Labs \\
  \texttt{\{sahibzada,jwhite,dhruv,g,uz\}@fastino.ai}}

\begin{document}
\maketitle

\begin{abstract}
Decision models score candidate answers in a single forward pass, but
the intermediate reasoning that improves decision quality adds latency.
We introduce \textbf{GLiDE}, a family of decision models from 0.4B to
27B parameters that distills reasoning into single-pass inference.
Building on Qwen's hybrid architecture, GLiDE removes causal masking
from full-attention layers so that context, question and candidate
answers interact bidirectionally. It is trained with soft-target
distillation from multiple teachers, including a reasoning model, and
progressive post-training on diverse decision tasks. On the Hugging Face \DI{}, each of the five GLiDE models
outscores every alternative of equal or smaller size. \dmtwo{} (27B)
ranks first on edition 0.4.1 (22~October 2026) with a hidden-set score
of 71.52, 2.0\% calibration error and 92.9\,ms median latency on one
GPU, and shares first place on all three versions of the Artificial
Analysis Decision Index. Both leaderboards are run by their maintainers
on closed test sets. With checkpoint averaging and recurrent depth,
GLiDE~27B reaches 73.37.
\end{abstract}

\begin{figure}[t]
\centering
\includegraphics[width=\columnwidth]{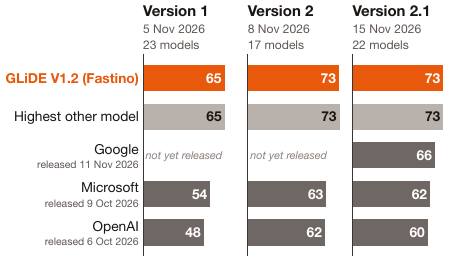}
\caption{\new{\textbf{\dmtwo{} remains within 0.6 points of the highest score across three versions of the Artificial Analysis Decision Index.} The model was not retrained between versions. Numbers are the scores the leaderboard showed on the date above each panel, in whole points. In unrounded scores \dmtwo{} is 0.2, 0.6 and 0.1 points below the highest model (Table~\ref{t:aa}). Scored by Artificial Analysis itself; part of its test data is private.}}
\label{f:aa_front}
\end{figure}

\section{Introduction}
\label{s:intro}

\paragraph{Decision models and the reasoning gap.}
Large language models are increasingly used not only to generate text, but also to make decisions: selecting tools, routing requests, classifying inputs, and choosing among predefined actions. In these settings, generating an answer token by token is often unnecessary. Decision models instead score candidate answers in a single forward pass, producing a probability distribution over the available choices. This approach has evolved from compact, schema-conditioned encoders such as GLiNER2~\citep{zaratiana2025gliner2} to larger systems including Jev~\citep{typesafe2026jev}, Decider~\citep{perplexity2026decider}, and Mammoth~\citep{baseten2026mammoth}. Yet a central tension remains: multi-step reasoning can substantially improve decision quality, while generating that reasoning introduces precisely the sequential computation that decision models seek to avoid. The challenge is therefore to capture the benefits of reasoning while retaining the speed of direct prediction, without sacrificing calibration or generalization~\citep{tian2023just,xiong2024can,recht2019imagenet}.

\begin{figure*}[t]
\centering
\includegraphics[width=\textwidth]{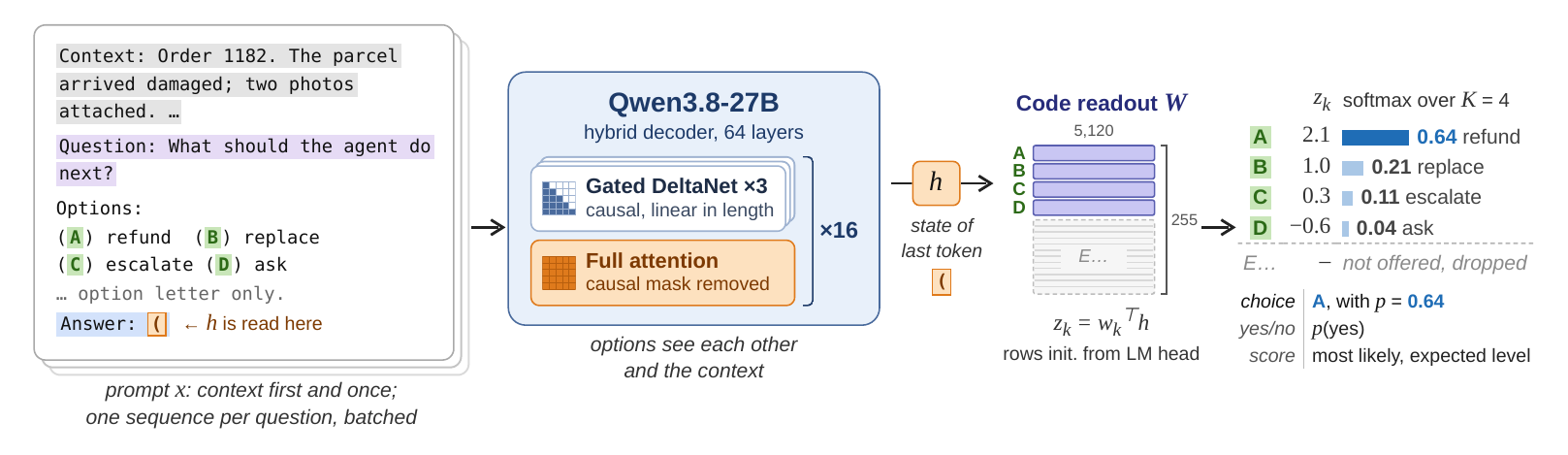}

\caption{\textbf{\dmtwo{} architecture.}
Each decision is encoded as a single sequence containing
the context, question, coded candidate answers, and an answer
prefix. Starting from a 64-layer Qwen hybrid decoder, GLiDE
removes causal masking from its 16 full-attention layers,
enabling bidirectional interactions while retaining causal
recurrence in the 48 Gated DeltaNet layers.
A dedicated 255-way readout scores the final hidden state,
and a softmax over the offered options produces a normalized
probability distribution. Decisions are computed in a single
forward pass, without generating tokens or intermediate
reasoning. Appendix~\ref{a:primer} details the output
semantics and deployment policies.}

\label{f:arch}
\end{figure*}

\paragraph{Our approach.}
We introduce \textbf{GLiDE}, a family of single-pass decision models spanning 0.4B to 27B parameters. Starting from Qwen hybrid language models, we remove causal masking from their full-attention layers, allowing the context, question, and candidate answers to interact bidirectionally while retaining causal Gated DeltaNet layers (Figure~\ref{f:arch}). A dedicated readout then produces probabilities over the offered options. To transfer reasoning-informed supervision into this architecture, we combine soft-target distillation from multiple teachers, including a model evaluated after reasoning, with progressive post-training on diverse decision tasks. In this way, GLiDE learns from the distributions produced by reasoning without generating intermediate reasoning at inference time.

\paragraph{Decision performance and calibration.}
We first evaluate GLiDE on the Hugging Face \DI{}~\citep{passos2026decisionindex}, which ranks models using independently administered hidden test sets. On edition~0.4.1 (22~October 2026), \dmtwo{} (27B) achieves a score of 71.52, ranking first among 225 models and outperforming the runner-up by 1.77 points (Figure~\ref{f:boardviz}). It also leads on both hidden components, covering unseen questions from established skills and decisions in new domains. This performance is accompanied by an expected calibration error of 2.0\% and a median latency of 92.9\,ms on a single RTX PRO 6000 GPU, demonstrating that strong decision accuracy can coexist with reliable confidence estimates and efficient inference.

\paragraph{Hidden-set evaluation.}
The Hugging Face Decision Index evaluates submitted models using private test sets administered by the benchmark maintainers. As of November 15, 2026, GLiDE models achieved competitive performance across multiple leaderboard revisions, with GLiDE 27B scoring 73.37 on the hidden evaluation sets. All five model sizes also lie on the reported performance--parameter Pareto frontier.

\paragraph{Scaling and recurrent depth.}
Finally, we investigate how the approach extends across model sizes and inference-time computation budgets. We develop four smaller GLiDE models through progressive training and distillation from the 27B model. All five models lie on the performance--parameter Pareto frontier of the Hugging Face \DI{}, each outperforming every evaluated alternative of equal or smaller size. We also investigate parameter-free recurrent depth, which increases computation by reapplying selected middle layers without introducing new weights or generating tokens. Combined with further checkpoint averaging, this produces GLiDE~27B, which reaches 73.37 on the Hugging Face \DI{} with an expected calibration error of 2.1\%. Together, these results establish GLiDE as a family of fast, calibrated decision models that benefit from reasoning-informed training without requiring explicit reasoning generation at inference time.

\noindent \paragraph{} Taken together, these findings show that strong, calibrated decision performance can be achieved across model scales without generating explicit reasoning at inference time. More broadly, GLiDE demonstrates how reasoning-informed supervision and additional computation within the model can improve decision quality while preserving the efficiency of direct candidate scoring.

\section{Decision Model Architecture}
\label{sec:architecture}

\new{Readers new to decision models may prefer to start with Appendix~\ref{a:walkthrough}, which builds the model from the base language model step by step.}

\subsection{Task Formulation}

We consider decision-making tasks in which a model selects
an answer from a finite set of candidates. A request
consists of a context $c$ and $M$ questions:
\begin{equation}
    \mathcal{R}
    = \left(c,\{(q_j,\mathcal{A}_j)\}_{j=1}^{M}\right),
\end{equation}
where $\mathcal{A}_j=\{a_{j,1},\ldots,a_{j,K_j}\}$
denotes the candidate answers for question $j$.
For brevity, let $u_j=(c,q_j,\mathcal{A}_j)$.
The model predicts a categorical distribution:
\begin{equation}
    p_\theta(\cdot\mid u_j)\in\Delta^{K_j-1},
    \qquad
    \sum_{k=1}^{K_j}p_\theta(k\mid u_j)=1.
\end{equation}
The predicted decision and its confidence are
\begin{equation}
    \begin{aligned}
        \hat{k}_j
        &=\arg\max_{k\in[K_j]}p_\theta(k\mid u_j),\\
        \hat{p}_j
        &=p_\theta(\hat{k}_j\mid u_j).
    \end{aligned}
\end{equation}
Unlike autoregressive generation, decision inference
requires a single forward pass, without generating
intermediate reasoning or answer tokens.
Appendix~\ref{a:primer} formalizes typed outputs,
cost-sensitive decisions, confidence-based deferral,
and the semantics of batched questions.

\subsection{Model Architecture}

Figure~\ref{f:arch} illustrates the architecture of
\dmtwo{}, from input encoding to single-pass decision
probabilities.

\paragraph{Input encoding.}
Each question is converted into a single sequence that
concatenates the context, question, and candidate answers.
The context appears once in each sequence,
while each candidate is assigned a unique single-token
code $\gamma_k$. A final answer prefix marks the position
from which the model predicts the decision distribution.
Formally, the input sequence is defined as

\begin{equation}
    \begin{aligned}
        x_j &= \operatorname{Tok}(c\oplus q_j\\
            &\quad\oplus
            \bigoplus_{k=1}^{K_j}
            (\gamma_k\oplus a_{j,k})
            \oplus s_{\mathrm{ans}}),
    \end{aligned}
\end{equation}
where $\oplus$ denotes concatenation and
$s_{\mathrm{ans}}$ is the final answer prefix.
Questions from the same request are processed
as separate sequences in one batch.

\paragraph{Hybrid decoder.}
We build on Qwen3.8-27B~\citep{qwen2026qwen38_27b}, which contains 48 Gated
DeltaNet layers~\citep{yang2025gated} and 16 full-attention layers.
Standard causal attention is defined by
\begin{equation}
    \operatorname{Attn}(Q,K,V;\Omega)
    =
    \operatorname{softmax}\!\left(
    \frac{QK^\top}{\sqrt{d_k}}+\Omega
    \right)V,
\end{equation}
with the causal mask
\begin{equation}
    \Omega_{ij}=
    \begin{cases}
        0, & j\leq i,\\
        -\infty, & j>i.
    \end{cases}
\end{equation}
We remove this mask from the 16 full-attention layers,
setting $\Omega=0$. This allows bidirectional interaction
between the context, question, and candidate answers.
The Gated DeltaNet layers retain their causal
recurrence and linear complexity in sequence length.

\paragraph{Option readout.}
Let $F_\theta$ denote the hybrid decoder and
$L_j$ the length of $x_j$. We extract the final
hidden state at the answer prefix:
\begin{equation}
    H_j=F_\theta(x_j),
    \qquad
    h_j=H_j[L_j]\in\mathbb{R}^{d}.
\end{equation}
A dedicated linear readout maps this representation
to logits over 255 possible option codes:
\begin{equation}
    z_j=Wh_j,
    \qquad
    W\in\mathbb{R}^{255\times d}.
\end{equation}
Since each question may have a different number of
candidate answers, we restrict the readout to the
$K_j$ option codes present in the input, discarding
logits associated with unused codes. A softmax over
the remaining logits then produces a normalized
probability distribution over the offered candidates:
\begin{equation}
    p_\theta(k\mid u_j)=
    \frac{\exp(z_{j,k})}
{\sum_{\ell=1}^{K_j}\exp(z_{j,\ell})}
    \label{eq:decision}
\end{equation}
for $k\in[K_j]$. The readout is initialized from
the corresponding rows of the pretrained language-model
head. Thus, a single evaluation of $F_\theta$
produces a normalized distribution over all offered
answers, without autoregressive decoding.

\section{Training}
\label{sec:training}

We train \dmtwo{} through progressive post-training,
transferring reasoning capabilities into single-pass
inference and refining them with diverse decision
data and multi-teacher supervision. Figure~\ref{f:pipe}
illustrates the evolution of the model across stages.

\subsection{Progressive Training and Distillation}

We train GLiDE through four successive stages, each
building on the checkpoint produced by the preceding
stage. The procedure first develops a reasoning-capable
teacher, then converts it into a single-pass decision
model before progressively refining its predictions.

\paragraph{Stage 1: Reasoning teacher.}
We first adapt \basemodel{} using LoRA~\citep{hu2022lora}
in two steps. We begin by generating reasoning traces
for training questions and retaining approximately
6,400 examples whose reasoning leads to the correct
answer. Rank-64 LoRA adapters are trained on these
examples with the base weights frozen, using
next-token prediction on the reasoning traces and
cross-entropy on the final answers. Next, the adapted model reasons over approximately
135K additional questions, and we collect its
post-thinking probability distributions over the
candidate answers. A second LoRA round trains the
model to predict these distributions directly, without
producing written reasoning. The resulting teacher
can operate with or without explicit reasoning,
providing both direct and post-thinking predictions
for subsequent distillation. This is the only stage
that uses LoRA.

\paragraph{Stage 2: Single-pass conversion.}
Starting from the reasoning teacher, we remove causal
masking from the full-attention layers and introduce
the dedicated option readout. We then distill soft
probability distributions from multiple teachers,
including the reasoning model's post-thinking
predictions, to train the architecture for direct
decision scoring in a single forward pass.

\paragraph{Stage 3: Short training rounds.}
We then continue training through five short rounds.
After each round, the updated checkpoint is averaged
equally with its parent, yielding \dmone{}.
This recursive averaging gives the latest update
half of the final weight, the preceding update
one quarter, and progressively smaller contributions
to earlier rounds, helping retain previously learned
capabilities while incorporating new refinements.

\paragraph{Stage 4: Final post-training.}
Finally, we continue training \dmone{} on a rebuilt
mixture of diverse decision tasks, using task-specific
multi-teacher distributions as supervision. This
stage consolidates the preceding improvements
and produces \dmtwo{}, our final decision model.

\paragraph{Distilling reasoning.}
The central challenge is to preserve the benefits
of slow, multi-step reasoning without performing
it at inference time. For input $x$, let
$q_{\mathrm{think}}$ and $q_{\mathrm{direct}}$
denote the reasoning teacher's answer distributions
with and without thinking, respectively. We construct the target distribution as
\begin{equation}
q =
\begin{cases}
\frac{1}{2}(q_{\mathrm{think}}+
q_{\mathrm{direct}}), & \kappa(x)<\eta,\\
q_{\mathrm{direct}}, & \kappa(x)\geq\eta,
\end{cases}
\label{eq:reason_target}
\end{equation}
where $\kappa(x)=\max_k q_{\mathrm{direct}}(k\mid x)$ is the direct model's confidence
and $\eta=0.8$. \new{In words: if the teacher is already at least 80\% confident without thinking, we use its direct distribution. Otherwise, we average its direct and post-thinking distributions.} Averaging the distributions retains the direct model's signal while incorporating changes induced by reasoning. This provides post-thinking
supervision primarily when the direct model
is uncertain. Given a gold answer $y$ and teacher distribution
$q$, the student minimizes
\begin{equation}
\begin{aligned}
\mathcal{L}(\theta)
={}&-\alpha\log p_\theta(y\mid x)\\
&+(1-\alpha)
D_{\mathrm{KL}}(q\|p_\theta),
\end{aligned}
\label{eq:training_loss}
\end{equation}
with $\alpha=0.3$. \new{The first term rewards the correct answer. The second makes the student match the teacher's full distribution.} Soft targets convey not only
the preferred answer but also the uncertainty
remaining after reasoning, allowing the student
to approximate the teacher's decision distribution
in a single pass.

\begin{table*}[]
\centering
\small
\setlength{\tabcolsep}{9pt}
\begin{tabular}{@{}lrrrrrr@{}}
\toprule
\textbf{Model} &
\textbf{Full} $\uparrow$ &
\textbf{Same-skill} $\uparrow$ &
\textbf{New-domain} $\uparrow$ &
\textbf{Gen. gap} &
\textbf{ECE (\%)} $\downarrow$ &
\textbf{Latency (ms)} $\downarrow$ \\
\midrule
\textbf{\dmtwo{}}
& \textbf{71.52} & \textbf{68.31}
& \textbf{76.32} & $-0.47$
& \textbf{2.0} & \textbf{92.9} \\
Jeffan-31B
& 69.75 & 65.86 & 75.58 & $-3.38$
& 4.7 & 121.0 \\
mammoth-1
& 68.44 & 65.98 & 72.14 & $-6.90$
& 4.1 & 220.1 \\
Decider v1.2 
& 67.98 & 63.99 & 73.98 & $+0.24$
& 5.5 & 144.4 \\
d3-turbo v1.1 
& 67.31 & 63.98 & 72.31 & $-4.31$
& 5.2 & 190.5 \\
\midrule
Jev 1.14
& 60.24 & 58.91 & 62.24 & $-0.46$
& 8.0 & 196.0$^\dagger$ \\
\bottomrule
\end{tabular}
\caption{\textbf{Decision Index comparison}
(edition 0.4.1, 22 October 2026).
Full is the hidden-set ranking score; Same-skill
and New-domain are its components.
Gen. gap is Full minus public score; closer to zero means that public and hidden scores agree.
ECE measures calibration; latency is measured
on one RTX PRO 6000 GPU
($^\dagger$hosted API).}
\label{t:main_results}
\end{table*}

\subsection{Training Data and Generation}

GLiDE is trained on a mixture of existing datasets
and generated decision examples spanning classification,
extraction, knowledge, reasoning, tool use, and safety.
We use procedural generation and language-model-based
annotation to expand task coverage and introduce
variation in question structure, candidate answers,
and difficulty. The final post-training stage contains 374,720
examples. Across the training pipeline, we use
more than 1.6 million distinct questions from over
500 task families, corresponding to approximately
2.3 billion training tokens. Table~\ref{t:data}
summarizes the scale of the training data.

\begin{table}[t]
\centering
\small
\setlength{\tabcolsep}{5pt}
\renewcommand{\arraystretch}{1.08}
\begin{tabularx}{\columnwidth}{
@{}>{\raggedright\arraybackslash}Xr@{}
}
\toprule
\textbf{Training characteristic} & \textbf{Value} \\
\midrule
Final-stage examples       & 374,720 \\
Distinct questions         & $>$1.6M \\
Training tokens            & $\sim$2.3B \\
Task families              & $>$500 \\
Screened domains           & 267 \\
\bottomrule
\end{tabularx}
\caption{\textbf{GLiDE training data overview.}
Training combines existing datasets, procedural
generation, and LLM-generated decision examples.
Totals cover all training stages unless otherwise
specified.}
\label{t:data}
\end{table}

\subsection{Multi-Teacher Supervision}

The final training stage combines complementary
teachers, including existing single-pass decision
models, an earlier checkpoint of our model, and
an annotator. Since teacher performance differs
across tasks, we use task-specific log-linear
pooling~\citep{genest1986combining}. For task family $f$, the aggregated target is
\begin{equation}
q_f(k\mid x) =
\frac{\exp(s_{f,k}(x))}
{\sum_{\ell=1}^{K_x}\exp(s_{f,\ell}(x))},
\label{eq:teacher_pool}
\end{equation}
where
\begin{equation}
s_{f,k}(x)
=\sum_{t=1}^{T}
\lambda_{f,t}\log p_t(k\mid x).
\end{equation}
Here $p_t$ denotes teacher $t$ and
$\lambda_{f,t}\in[0,4]$ its task-specific weight.
We optimize these weights by minimizing gold-label
negative log-likelihood on a held-out 5\% slice of the training data, using two-fold cross-fitting.

\paragraph{Gold anchoring.}
To limit propagation of teacher errors, pooled
targets are subject to gold-consistency checks and
replaced by gold-anchored targets when necessary.
The downstream student gain from this procedure
remains within single-seed variability.

\paragraph{Abstention.}
For knowledge tasks, ``I don't know'' is treated
as an ordinary candidate answer. Rather than
imitating teacher uncertainty, we derive
abstention targets from the starting student's
confidence, encouraging it to abstain when
its own knowledge is insufficient.

\subsection{Optimization Details}

The final stage uses full-parameter fine-tuning
with AdamW~\citep{loshchilov2019decoupled},
a learning rate of $10^{-6}$, batch size 128,
and 2,928 steps over one epoch. Candidate
options are shuffled during training to
reduce position bias. Each seed requires
approximately six hours on eight B200 GPUs.
The released model averages the weights of
two independently trained seeds
~\citep{wortsman2022soups}.

\section{Experimental Setup}
\label{sec:experiments}

We evaluate \dmtwo{} on decision-making benchmarks
covering accuracy, generalization, calibration,
and inference efficiency. Our primary evaluation
is the Hugging Face \DI{}, complemented by
externally administered benchmarks and held-out
internal evaluation sets.

\begin{figure}[]
\centering
\includegraphics[width=0.99\columnwidth]{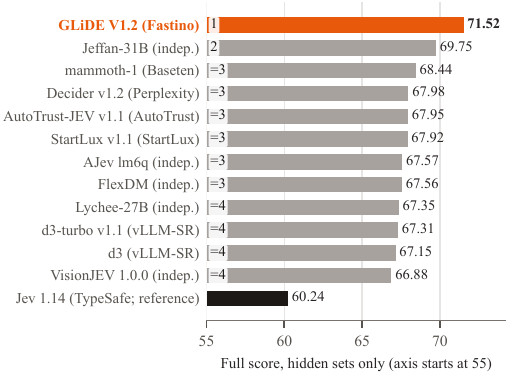}
\caption{\textbf{Hugging Face \DI{}, edition 0.4.1}
(22 October 2026; \citealp{passos2026decisionindex}).
Full score on the hidden sets for the top 12 of 225 models,
plus Jev~1.14~\citep{typesafe2026jev} as reference.
=3 and =4 denote ranks shared under the board's tie rule;
``indep.'' denotes an independent developer.
The axis starts at 55. Scored by the board's maintainers on hidden test sets.}
\label{f:board}
\end{figure}

\subsection{Benchmarks}
\label{sec:benchmarks}

\paragraph{Decision Index.}
The \DI{}~\citep{passos2026decisionindex} evaluates
decision models across 37 public benchmarks spanning
five task categories. Since October 20, 2026,
its official ranking has relied exclusively on two
hidden evaluation sets administered by the
benchmark maintainers:
(i)~\emph{same-skill}, measuring generalization to
unseen questions from known task families, and
(ii)~\emph{new-domain}, evaluating decisions in
domains absent from the public benchmarks. Let $S$ and $O$ denote the normalized scores on
the two hidden sets. The overall ranking score is $\mathrm{Full}=0.6S+0.4O$.
Models must complete inference without intermediate
reasoning and satisfy a one-second latency limit
on a single RTX PRO 6000 GPU. We report results
from edition 0.4.1 (October 22, 2026).

\new{
\paragraph{Artificial Analysis Decision Index.}
We also evaluate GLiDE on the Artificial Analysis
Decision Index~\citep{aa2026decisionindex}, an
externally administered benchmark assessing
decision-making capabilities across knowledge,
reasoning, tool use, and safety. Artificial Analysis
evaluates models through their public APIs using a
standardized protocol and publishes the resulting
scores independently of model developers. The original version combines four equally weighted
evaluations: AA-Omniscience-Decision, which measures
knowledge reliability and selective answering
\citep{jackson2025omniscience}; multiple-choice
questions from Humanity's Last Exam
\citep{cais2026hle}; When2Call, which evaluates
tool-use decisions~\citep{ross2025when2call};
and WildGuardTest, which assesses safety
classification~\citep{han2024wildguard}.
We report the official scores published by
Artificial Analysis.
} \new{The index has been revised twice since. Version~2 (8~November 2026; \citealp{aa2026decisionindexv2}) uses six evaluations. It adds decisions about documents and a test of whether a decision stays the same when the question is reworded, and it draws 45\% of its weight from private data, up from 25\%. Version~2.1 followed within days. It revised the consistency test and rescored every model. We report all three versions.}

\new{\paragraph{Hidden test sets.}
The Hugging Face ranking uses only hidden questions. The Artificial
Analysis index combines a private set with public ones. We have never
had access to the hidden questions, their answers, or per-question
results of either benchmark; we see only the aggregate scores
published by their maintainers.}

\paragraph{Complementary evaluations.}
To assess performance beyond the primary leaderboard,
we also evaluate on JevBench,
SciFact, a held-out clinical decision set,
and an internal regression suite spanning eight
real-world decision categories.

\begin{table}[t]
\centering
\small
\setlength{\tabcolsep}{4pt}
\begin{tabular}{@{}lrrr@{}}
\toprule
\textbf{Configuration} &
\textbf{Dev.} $\uparrow$ &
\textbf{Acc.} $\uparrow$ &
\textbf{Sens.} $\downarrow$ \\
\midrule
Stage 1 (no training) & 59.36 & -- & 1.36 \\
Causal + gold labels & 60.42 & 87.01 & 1.25 \\
Non-causal + gold & 61.78 & 88.38 & 0.22 \\
Non-causal + soft targets & \textbf{62.59}
& \textbf{89.16} & $-0.18$ \\
+ 30\% format augmentation & 62.22
& 88.67 & -- \\
\bottomrule
\end{tabular}
\caption{\textbf{Single-pass conversion ablations.}
All trained variants use 125K examples and
one seed. Dev. is a composite development score;
Acc. is held-out accuracy (\%); Sens. measures
score loss under input-format perturbations.
Negative Sens. means that the score improved under perturbation; -- denotes not measured.}
\label{t:architecture_ablation}
\end{table}

\subsection{Evaluation Metrics}

We evaluate three complementary properties of decision models: predictive accuracy, calibration, and inference efficiency.

\paragraph{Decision performance.}
We report the official benchmark scores and
the two hidden components $S$ and $O$.
For the \DI{}, public benchmarks are scored
using chance-corrected skill and do not
contribute to the official ranking.

\paragraph{Calibration.}
We measure expected calibration error (ECE),
which compares predicted confidence with
empirical accuracy across $B$ confidence bins:
\begin{equation}
    \mathrm{ECE}
    =\sum_{b=1}^{B}
    \frac{n_b}{N}
    \left|\mathrm{acc}(b)-\mathrm{conf}(b)\right|,
    \label{eq:ece}
\end{equation}
where $n_b$ is the number of predictions in
bin $b$. Lower ECE indicates better calibration.

\paragraph{Inference efficiency.}
We report median inference latency on a single
RTX PRO 6000 GPU, following the leaderboard
protocol. All decisions are produced in one
forward pass, without autoregressive decoding.

\subsection{Development Evaluation}
\label{sec:selection}

We evaluate candidate checkpoints on internal
development sets covering accuracy on held-out task families and domains, input robustness, and regression checks.
Final benchmark scores are obtained through
evaluations administered by the benchmark
maintainers. Appendix~\ref{a:sel} provides
additional methodological context.

\begin{table}[t]
\centering
\normalsize
\setlength{\tabcolsep}{5pt}
\renewcommand{\arraystretch}{1.05}
\resizebox{\linewidth}{!}{%
\begin{tabular}{@{}lrrr@{}}
\toprule
\textbf{Benchmark} &
\textbf{\dmtwo{}} &
\textbf{Best open} &
\textbf{Jev 1.13} \\
\midrule
JevBench (accuracy) & 93.07 & \textbf{93.94} & 86.58 \\
Clinical held-out & \textbf{97.01} & 96.56 & -- \\
SciFact (macro-F1) & \textbf{85.63} & 84.05 & -- \\
Regression watch & \textbf{84.88} & 84.10 & -- \\
\bottomrule
\end{tabular}%
}
\caption{\textbf{Complementary development measurements}}
\label{t:external}
\end{table}

\section{Results}
\label{s:results}

We evaluate \dmtwo{} in terms of decision accuracy,
generalization, calibration, and inference efficiency.
We further examine the contributions of non-causal
attention and soft-target distillation.
\new{Each paragraph below opens with the question it answers.}

\begin{figure*}[t]
\centering
\includegraphics[width=\textwidth]{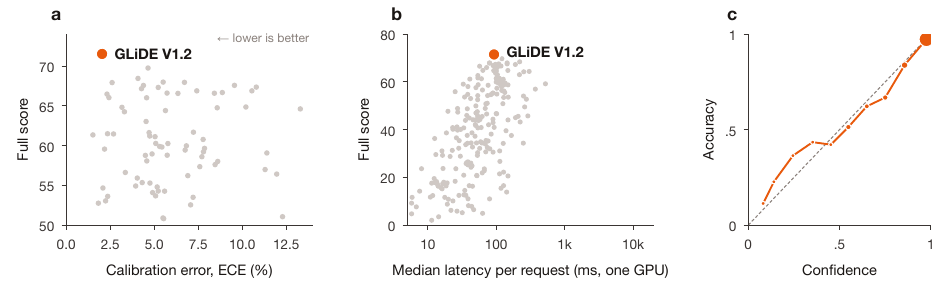}
\caption{\new{\textbf{\dmtwo{} compared with the other models on the Hugging Face \DI{}, edition 0.4.1.} Scored by the board's maintainers on hidden test sets. \textbf{a},~Calibration error against Full score, for models scoring above 50. \textbf{b},~Full score against median latency on one GPU. \textbf{c},~Stated confidence against measured accuracy for \dmtwo{}; larger markers hold more answers.}}
\label{f:boardviz}
\end{figure*}

\subsection{Main Results}

\paragraph{Decision performance.}
\new{\emph{Does \dmtwo{} decide better than other models on questions it has never seen?}}
Table~\ref{t:main_results} reports results on the Hugging Face \DI{} (edition 0.4.1, 22~October 2026). \dmtwo{} achieves a Full score of 71.52, ranking first of 225 models, 1.77 points ahead of Jeffan-31B; this margin is about twice the board's 0.9-point tie margin, while six models share third place within 0.9 points of one another. \dmtwo{} is also first on each hidden component: 68.31 on same-skill (next best 65.98) and 76.32 on new-domain (next best 75.58).
\new{Figure~\ref{f:boardviz} shows that this lead does not come at the cost of calibration or speed. Every model with better calibration scores at most 61.3, and every faster model scores at most 68.0.} \new{A comparison with Drex Max~\citep{dhankhar2026drex} helps isolate the effect of training. Drex Max starts from the same \basemodel{} base and makes the same two architectural changes: non-causal attention and a 255-way readout. It scores 66.60 on this board, 4.92 points below \dmtwo{}. The two models share a backbone and readout but differ in how they were trained.}

\paragraph{Generalization.}
\new{\emph{Does the model do as well on hidden questions as on public ones?}}
\dmtwo{}'s Full score is 0.47 points below its public score. Across the 225 models, Full scores track public ones (slope 0.98, Spearman $\rho=0.95$) but sit about 4 points lower; a model with \dmtwo{}'s public score is expected to reach 67.3, and \dmtwo{}'s 71.52 lies inside the 95\% prediction interval [55.0, 79.5] of a fit that excludes it. The small gap is consistent with good generalization, although a single benchmark cannot establish distribution-wide robustness.

\new{\paragraph{Artificial Analysis evaluation.}
We additionally evaluate the frozen \dmtwo{}
checkpoint on three successive versions of the
Artificial Analysis Decision Index, administered
through the model's public API. Across these
versions, GLiDE remains within 0.6 points of
the highest unrounded score. The subsequent revisions introduce new tasks
and provide complementary evidence of performance
stability under changes to the benchmark.
Table~\ref{t:aa} reports the results, and
Appendix~\ref{a:aa} lists all evaluated models.}

\begin{table}[t]
\centering\footnotesize\setlength{\tabcolsep}{2.4pt}
\new{\begin{tabular}{@{}llrrr@{}}
\toprule
Model & Developer & v1 & v2 & v2.1 \\
\midrule
Drex Max v1.1 & Nace.AI & 63.78 & \underline{73.39} & \underline{72.62} \\
SubQ Snap 1.2 & Subquadratic & -- & -- & \underline{72.58} \\
\textbf{GLiDE V1.2} & Fastino & \underline{65.01} & \underline{72.83} & \underline{72.52} \\
Mammoth 27B & Baseten & -- & 72.45 & 71.36 \\
d1-ultra & Liquid AI & \underline{65.25} & 70.45 & 70.58 \\
SubQ Snap 1.1 & Subquadratic & \underline{65.00} & 69.17 & 67.92 \\
Elephant 9B & Baseten & \underline{64.69} & 67.55 & 66.39 \\
Jev 1.14 & TypeSafe & 57.07 & 66.96 & 66.23 \\
Gemini Decision 1 & Google & -- & -- & 65.98 \\
Microsoft-Decision-1 & Microsoft & 53.76 & 62.80 & 62.07 \\
GPT-6 Luna (Decisions) & OpenAI & 48.49 & 61.73 & 60.25 \\
\midrule
Highest score & & 65.25 & 73.39 & 72.62 \\
Highest displayed score & & 65 & 73 & 73 \\
Models listed & & 23 & 17 & 22 \\
\bottomrule
\end{tabular}}
\caption{\new{\textbf{Artificial Analysis Decision Index, versions 1, 2 and 2.1.} Scores are those published by Artificial Analysis, retrieved on 5, 8 and 15~November 2026. The leaderboard displays them as whole numbers. An underlined score displays as the highest number on its version. Rows are ordered by version~2.1 score; a dash means the model was not listed on that version. Shown are every model within one point of the highest score on any version, plus Jev and the models of Google, Microsoft and OpenAI. Appendix~\ref{a:aa} lists all models. Higher is better.}}
\label{t:aa}
\end{table}

\subsection{Calibration and Efficiency}

\paragraph{Probability calibration.}
\new{\emph{When the model says it is 90\% sure, is it right about 90\% of the time?}}
\dmtwo{} has an expected calibration error of 2.0\% (10 bins on the board's sample; interval about [1.7, 2.4]), the lowest among the ten highest-ranked models (Table~\ref{t:cal}). Its mean confidence exceeds its accuracy by 1.1 points (82.6\% vs.\ 81.5\%), whereas the other nine are overconfident by 2.5 to 10.5 points. The 56\% of its answers given with confidence above 0.9 are correct 97.4\% of the time. Its Brier score, which mixes calibration and accuracy, is slightly higher than those of three models that are 0.9--1.3 points more accurate on this sample.
\new{We apply no correction after training, such as a fitted temperature. The probabilities are the plain softmax of the readout, so the calibration is a result of training and not of a later fix. Calibration differs by task area: ECE is 1.8\% on language tasks, 3.5\% on retrieval, 3.9\% on tools, 4.2\% on arts and 7.5\% on knowledge. Figure~\ref{f:boardviz}c plots confidence against accuracy.}

\new{\paragraph{Acting only on confident answers.}
\emph{If an application acts only when the model is confident, how accurate are those answers?}
Table~\ref{t:coverage} answers this on the same sample. Accepting every answer gives 81.5\% accuracy. Accepting only answers with a confidence of at least 0.8 keeps 68\% of them, and 95.0\% of those are correct. At 0.9, 56\% are kept and 97.4\% are correct. An application can therefore trade coverage for accuracy by setting a threshold, with no retraining (Appendix~\ref{a:primer}).}

\begin{table}[t]
\centering\footnotesize\setlength{\tabcolsep}{5pt}
\new{\begin{tabular}{@{}lrr@{}}
\toprule
Confidence at least & Answers kept (\%) & Accuracy (\%) \\
\midrule
0 (all answers) & 100.0 & 81.5 \\
0.5 & 90.8 & 86.2 \\
0.6 & 82.9 & 89.5 \\
0.7 & 76.1 & 92.0 \\
0.8 & 67.9 & 95.0 \\
0.9 & 56.0 & 97.4 \\
\bottomrule
\end{tabular}}
\caption{\new{\textbf{Accuracy when only confident answers are accepted.} Computed for \dmtwo{} from the board's calibration sample.}}
\label{t:coverage}
\end{table}

\paragraph{Inference latency.}
\new{\emph{How fast is the model?}}
\dmtwo{} has a median latency of 92.9\,ms on a single RTX PRO 6000 GPU, $1.3\times$ faster than Jeffan-31B, $2.4\times$ faster than mammoth-1 and $2.1\times$ faster than d3-turbo v1.1; every faster model on the board scores at least 3.5 points lower.
\new{Figure~\ref{f:boardviz}b shows score against latency for all models.}

\paragraph{Complementary evaluations.}
\new{\emph{How does the model perform on additional task sets?}}
\dmtwo{} is best on three of four complementary benchmarks in Table~\ref{t:external}; on JevBench it trails the best open model by 0.87 points.

\begin{table}[t]
\centering\footnotesize\setlength{\tabcolsep}{2pt}
\begin{tabular}{@{}lrrrr@{}}
\toprule
Model & ECE & Conf.$-$acc. & Wrong, conf.\,${\geq}\,$0.95 & Brier \\
\midrule
\dmtwo{} & \textbf{2.0} & \textbf{+1.1} & 0.72 & 0.257 \\
StartLux v1.1 & 2.6 & +2.5 & 1.28 & 0.239 \\
mammoth-1 & 4.1 & +3.9 & 1.37 & \textbf{0.238} \\
Jeffan-31B & 4.7 & +4.7 & 1.64 & 0.249 \\
FlexDM & 4.4 & +3.5 & \textbf{0.25} & 0.355 \\
Jev 1.14 & 8.0 & +7.3 & 2.65 & 0.357 \\
\bottomrule
\end{tabular}
\caption{\textbf{Calibration on the board's sample} Lower is better, except for Conf.$-$acc., where closer to zero is better; a positive value means overconfidence. \new{ECE and the fourth column are in \%. The fourth column is the share of answers that are wrong despite a confidence of at least 0.95.}}
\label{t:cal}
\end{table}

\begin{figure*}[t]
\centering
\includegraphics[width=\textwidth]{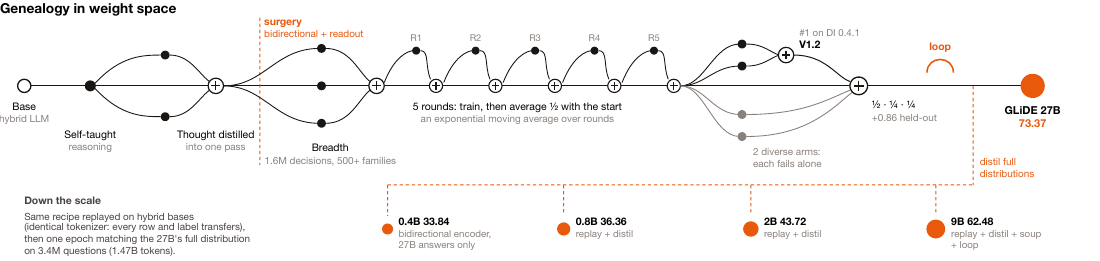}
\caption{\new{\textbf{Every GLiDE model descends from one lineage of weights.} Filled circles are fine-tuned checkpoints, $\oplus$ marks a weight average, and the arc marks the recurrent loop, which needs no training. Dashed lines show where the output distributions of the 27B model are distilled into smaller models. Scores are Full scores on the Hugging Face \DI{}, scored by the board's maintainers on hidden test sets. ``\#1 on DI 0.4.1'' refers to the board of 22~October 2026.}}
\label{f:genealogy}
\end{figure*}

\subsection{Ablation Study}

\new{\emph{Which part of the single-pass conversion helps: removing the causal mask, training on soft targets, or both?}
Table~\ref{t:architecture_ablation} changes one thing per row. The first row is the Stage~1 model before any conversion training. The second keeps the causal mask and trains on gold labels. The third removes the mask. The fourth replaces gold labels with soft targets. The last also re-formats 30\% of the examples. The table reports a composite development score, held-out accuracy, and sensitivity to input-format perturbations.}

Table~\ref{t:architecture_ablation} compares single-pass conversion variants trained on the same 125K examples with one seed. Removing the causal mask raises the development score from 60.42 to 61.78 and reduces format sensitivity from 1.25 to 0.22 points; adding soft targets raises the development score to 62.59 and held-out accuracy to 89.16\%. All three measures move in the same direction, but each step is within single-seed variability, so we treat these results as indicative rather than conclusive. Format augmentation did not help and was not retained.

\begin{figure*}[t]
\centering
\includegraphics[width=\textwidth]{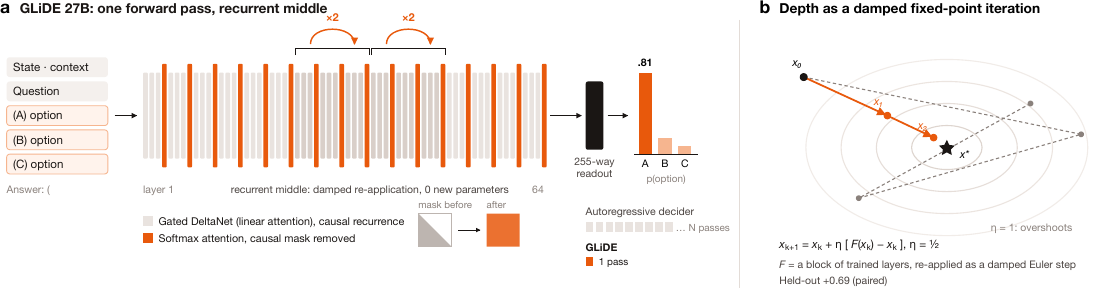}
\caption{\new{\textbf{Recurrent depth.} \textbf{a},~GLiDE~27B in one forward pass: the 64-layer hybrid decoder with the causal mask removed from its 16 full-attention layers, the 255-way readout, and two blocks of middle layers that are each applied twice at inference. \textbf{b},~The loop treats a block of layers $F$ as a map and takes damped steps towards its fixed point (Equation~\ref{eq:loop}). An undamped step ($\eta=1$) overshoots.}}
\label{f:loop}
\end{figure*}

\section{\new{Recurrent Depth and Model Scaling}}
\label{s:later}

\new{We investigate two extensions of \dmtwo{}: increasing inference-time computation without introducing additional parameters, and transferring the training procedure to smaller model scales. Figure~\ref{f:genealogy} summarizes the resulting model lineage. Unless stated otherwise, development uses internal evaluation data and reported comparisons use separate held-out sets. Hugging Face \DI{} scores are externally administered measurements from edition~0.4.1 ~\citep{passos2026decisionindexnov}.}

\subsection{\new{GLiDE 27B}}

\new{\paragraph{Parameter averaging.}
We train two additional checkpoints from the same lineage on broader mixtures of approximately 1.7~million questions each. Individually, they change held-out performance by $+0.21$ and $-0.20$ points and do not satisfy the predefined improvement threshold of $+0.5$. We then form a convex parameter average assigning weight $1/2$ to \dmtwo{} and $1/4$ to each additional checkpoint. The averaged model improves held-out performance by $0.86$ points. This result is consistent with prior observations that checkpoints from a shared fine-tuning lineage can remain connected in weight space~\citep{wortsman2022soups,izmailov2018averaging}. Across the complete lineage, training uses approximately 3.9~billion tokens from 3.4~million distinct questions.}

\new{\paragraph{Parameter-free recurrent depth.}
GLiDE~27B uses the parameters of the averaged model without further training. At inference time, two middle blocks---layers 24--35 and 36--47 of the 64-layer network---are each applied twice. For a block map $F$, the recurrent update is
\begin{equation}
    x_{k+1}=x_k+\eta\,[F(x_k)-x_k],
    \;\; k=0,1,
    \;\; \eta=\tfrac{1}{2},
    \label{eq:loop}
\end{equation}
where $x_0$ is the block input and $x_2$ is passed to the subsequent layer. Equation~\ref{eq:loop} defines a damped fixed-point iteration related to weight-tied and recurrent-depth networks~\citep{dehghani2019universal,bai2019deep,geiping2025scaling,saunshi2025reasoning}. It adds computation but no trainable parameters. The damping coefficient limits the distribution shift presented to downstream layers; in our experiments, the undamped update ($\eta=1$) performs worse. Figure~\ref{f:loop} illustrates the procedure.}

\new{\paragraph{Evaluation protocol.}
Candidate configurations vary the repeated layer blocks, recurrence count, and damping schedule. Configuration choice uses internal development data. We compare the selected configuration with the averaged-model baseline on two separate held-out sets.}

\new{\paragraph{Results.}
Recurrent depth improves the two held-out scores by $0.69$ points (interval $[0.02, 1.42]$) and $0.56$ points ($[0.27, 0.89]$), respectively. GLiDE~27B obtains a hidden-set score of 73.37, compared with 71.52 for \dmtwo{}; this difference includes both parameter averaging and recurrent depth. Median latency increases from 92.9 to 122\,ms because 24 of the 64 layers are evaluated twice. Calibration remains similar, with ECE increasing from 2.0\% to 2.1\%. On the benchmark calibration sample, 58\% of predictions have confidence above 0.9, of which 97.3\% are correct.}

\new{\paragraph{Ablations.}
Applying each block three times performs worse than applying it twice, and introducing trainable loop parameters does not improve over the parameter-free update. High-learning-rate distillation and alternative parameter averages also fail to improve held-out performance.}

\subsection{\new{Scaling Across Model Sizes}}

\new{\paragraph{Training procedure.}
We construct four smaller variants: GLiDE~9B, 2B, 0.8B, and 0.4B. The 9B, 2B, and 0.8B variants use base models with the same hybrid architecture and tokenizer as \basemodel{}, allowing the examples and option labels from Section~\ref{sec:training} to transfer directly. A final distillation epoch trains each model to match the complete output distribution of GLiDE~27B over 3.4~million questions. This step improves held-out performance by 2.4 points at 0.8B parameters and 1.1 points at 2B parameters. GLiDE~9B additionally uses low-learning-rate distillation, parameter averaging, and recurrent depth. GLiDE~0.4B uses a bidirectional encoder trained against the same 27B output distributions. Generating labels for the 3.4~million-question corpus required 4.5~hours on 22~GPUs.}

\new{\paragraph{Data screening.}
Before training, later data underwent repeated near-duplicate screening using n-gram overlap, MinHash similarity, and item-level rules. Matching questions were dropped and not replaced. Appendix~\ref{a:data_quality} describes the corresponding controls for the final-stage mixture.}

\new{\paragraph{Scaling results.}
Table~\ref{t:family} and Figure~\ref{f:family} report the externally administered measurements. All five GLiDE models lie on the score--parameter Pareto frontier and exceed every evaluated alternative of equal or smaller size by 3.2 to 10.9 points. Median latency ranges from 18 to 122\,ms. GLiDE~27B has the second-highest score among 300 models, while GLiDE~9B is the smallest listed model that exceeds the Jev~1.14 reference score. The public weights for all five models are available online.\footnote{\url{https://huggingface.co/fastino}}}

\begin{figure}[t]
\centering
\includegraphics[width=0.94\columnwidth]{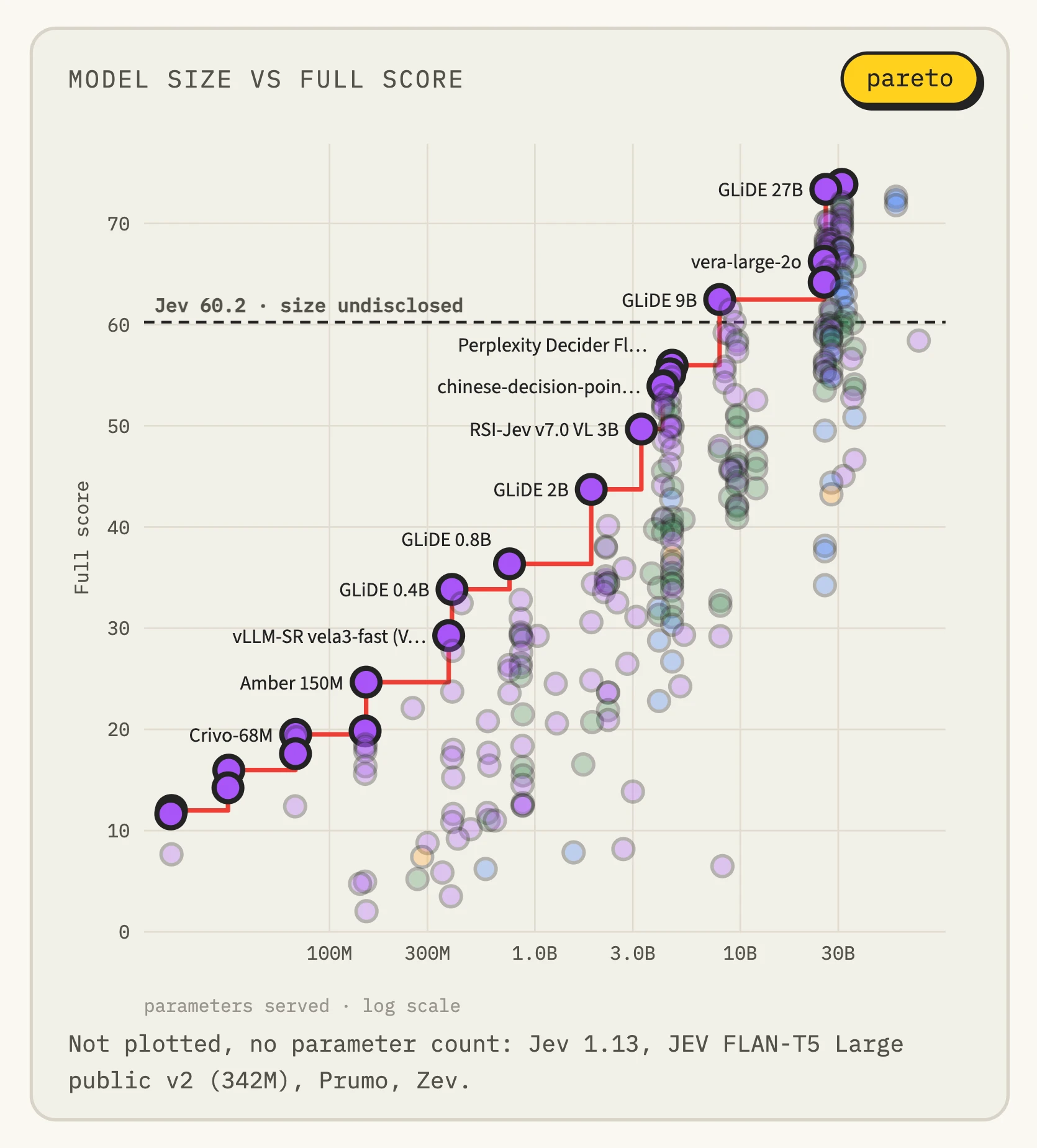}
\caption{\new{\textbf{The board's own chart of model size against Full score}, with its Pareto option switched on. Screenshot of the Hugging Face \DI{} website~\citep{passos2026decisionindexnov}, taken on 15~November 2026 and reproduced unchanged. Highlighted points are on the frontier. Figure~\ref{f:family}a redraws the same data. Scored by the board's maintainers on hidden test sets.}}
\label{f:screenshot}
\end{figure}

\begin{table}[t]
\centering\footnotesize\setlength{\tabcolsep}{3.5pt}
\new{\begin{tabular}{@{}lrrrrr@{}}
\toprule
Model & Params & Full & $\Delta$ & Latency & ECE \\
\midrule
GLiDE 27B & 26.1B & 73.37 & $+3.19$ & 122\,ms & 2.1\% \\
GLiDE 9B & 7.9B & 62.48 & $+6.49$ & 86\,ms & 1.7\% \\
GLiDE 2B & 1.9B & 43.72 & $+10.91$ & 37\,ms & 6.6\% \\
GLiDE 0.8B & 0.75B & 36.36 & $+3.89$ & 37\,ms & 7.6\% \\
GLiDE 0.4B & 0.39B & 33.84 & $+4.57$ & 18\,ms & 4.9\% \\
\bottomrule
\end{tabular}}
\caption{\new{\textbf{The GLiDE family on the Hugging Face \DI{}} (board of 12~November 2026, 300 models). Full is the ranking score on the hidden sets. $\Delta$ is the lead over the best other model of equal or smaller size. Latency is the board's median on one RTX PRO 6000 GPU.}}
\label{t:family}
\end{table}

\begin{figure*}[t]
\centering
\includegraphics[width=\textwidth]{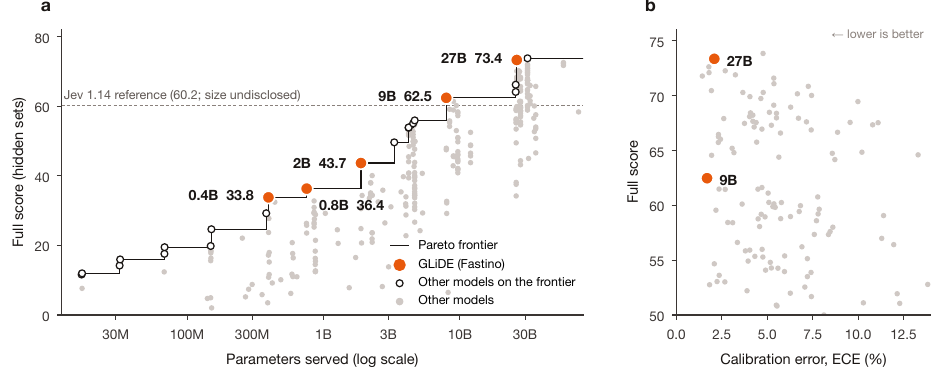}
\caption{\new{\textbf{The GLiDE family on the Hugging Face \DI{}} (board of 12~November 2026, 300 models). Scored by the board's maintainers on hidden test sets. \textbf{a},~Full score on the hidden sets against parameters served, for every model with a declared size. The line is the Pareto frontier: a model is on it when no smaller model scores higher. The five GLiDE models (orange) are on it; open circles are the other models on it. \textbf{b},~Calibration error against Full score, for models scoring above 50. Figure~\ref{f:screenshot} is the board's own version of panel~a.}}
\label{f:family}
\end{figure*}

\section{Conclusion}
\label{s:conc}

We introduced \dmtwo{}, a decision model that distills reasoning into a single forward pass. By combining non-causal full attention, a dedicated option readout, and multi-teacher distillation, the model produces calibrated probabilities over candidate decisions without generating intermediate reasoning. On the Hugging Face \DI{} (edition 0.4.1, 22~October 2026), \dmtwo{} achieves a score of 71.52, ranking first among 225 models and leading on both hidden evaluation sets, with an expected calibration error of 2.0\% and a median latency of 92.9\,ms. \new{These results extend beyond the initial model. Without retraining, \dmtwo{} remains within 0.6 points of the highest score across three versions of the Artificial Analysis Decision Index. Further checkpoint averaging and parameter-free recurrent depth yield GLiDE~27B, which reaches 73.37 without introducing additional parameters. Across five model sizes spanning 0.4B to 27B parameters, each GLiDE model outperforms all evaluated alternatives of equal or smaller size.} Together, these findings demonstrate that reasoning-informed training can enable strong, calibrated decision-making across model scales while preserving the efficiency of direct inference.

\bibliography{refs}

@inproceedings{yang2025gated,
  title = {Gated Delta Networks: Improving {Mamba2} with Delta Rule},
  author = {Yang, Songlin and Kautz, Jan and Hatamizadeh, Ali},
  booktitle = {The Thirteenth International Conference on Learning Representations},
  year = {2025},
  url = {https://openreview.net/forum?id=r8H7xhYPwz},
}

@inproceedings{behnamghader2024llm2vec,
  title = {{LLM2Vec}: Large Language Models Are Secretly Powerful Text Encoders},
  author = {BehnamGhader, Parishad and Adlakha, Vaibhav and Mosbach, Marius and Bahdanau, Dzmitry and Chapados, Nicolas and Reddy, Siva},
  booktitle = {First Conference on Language Modeling},
  year = {2024},
  url = {https://openreview.net/forum?id=IW1PR7vEBf},
}

@inproceedings{lee2025nvembed,
  title = {{NV-Embed}: Improved Techniques for Training {LLMs} as Generalist Embedding Models},
  author = {Lee, Chankyu and Roy, Rajarshi and Xu, Mengyao and Raiman, Jonathan and Shoeybi, Mohammad and Catanzaro, Bryan and Ping, Wei},
  booktitle = {The Thirteenth International Conference on Learning Representations},
  year = {2025},
  url = {https://openreview.net/forum?id=lgsyLSsDRe},
}

@article{hinton2015distilling,
  title = {Distilling the Knowledge in a Neural Network},
  author = {Hinton, Geoffrey and Vinyals, Oriol and Dean, Jeff},
  journal = {arXiv preprint arXiv:1503.02531},
  year = {2015},
  url = {https://arxiv.org/abs/1503.02531},
  note = {NIPS 2014 Deep Learning Workshop},
}

@inproceedings{kim2016sequence,
    title = "Sequence-Level Knowledge Distillation",
    author = "Kim, Yoon  and
      Rush, Alexander M.",
    editor = "Su, Jian  and
      Duh, Kevin  and
      Carreras, Xavier",
    booktitle = "Proceedings of the 2016 Conference on Empirical Methods in Natural Language Processing",
    month = nov,
    year = "2016",
    address = "Austin, Texas",
    publisher = "Association for Computational Linguistics",
    url = "https://aclanthology.org/D16-1139/",
    doi = "10.18653/v1/D16-1139",
    pages = "1317--1327"
}

@article{genest1986combining,
  title = {Combining Probability Distributions: A Critique and an Annotated Bibliography},
  author = {Genest, Christian and Zidek, James V.},
  journal = {Statistical Science},
  volume = {1},
  number = {1},
  pages = {114--135},
  year = {1986},
  publisher = {Institute of Mathematical Statistics},
  doi = {10.1214/ss/1177013825},
  url = {https://doi.org/10.1214/ss/1177013825},
}

@inproceedings{wortsman2022soups,
  title = 	 {Model soups: averaging weights of multiple fine-tuned models improves accuracy without increasing inference time},
  author =       {Wortsman, Mitchell and Ilharco, Gabriel and Gadre, Samir Yitzhak and Roelofs, Rebecca and Gontijo-Lopes, Raphael and Morcos, Ari S and Namkoong, Hongseok and Farhadi, Ali and Carmon, Yair and Kornblith, Simon and Schmidt, Ludwig},
  booktitle = 	 {Proceedings of the 39th International Conference on Machine Learning},
  pages = 	 {23965--23998},
  year = 	 {2022},
  editor = 	 {Chaudhuri, Kamalika and Jegelka, Stefanie and Song, Le and Szepesvari, Csaba and Niu, Gang and Sabato, Sivan},
  volume = 	 {162},
  series = 	 {Proceedings of Machine Learning Research},
  month = 	 {17--23 Jul},
  publisher =    {PMLR},
  url = 	 {https://proceedings.mlr.press/v162/wortsman22a.html},
}

@inproceedings{izmailov2018averaging,
  title = {Averaging Weights Leads to Wider Optima and Better Generalization},
  author = {Izmailov, Pavel and Podoprikhin, Dmitrii and Garipov, Timur and Vetrov, Dmitry and Wilson, Andrew Gordon},
  booktitle = {Proceedings of the Thirty-Fourth Conference on Uncertainty in Artificial Intelligence (UAI)},
  year = {2018},
  url = {https://arxiv.org/abs/1803.05407},
}

@article{chow1970optimum,
  title = {On Optimum Recognition Error and Reject Tradeoff},
  author = {Chow, C.},
  journal = {IEEE Transactions on Information Theory},
  volume = {16},
  number = {1},
  pages = {41--46},
  year = {1970},
  doi = {10.1109/TIT.1970.1054406},
  url = {https://doi.org/10.1109/TIT.1970.1054406},
}

@inproceedings{guo2017calibration,
  title = 	 {On Calibration of Modern Neural Networks},
  author =       {Chuan Guo and Geoff Pleiss and Yu Sun and Kilian Q. Weinberger},
  booktitle = 	 {Proceedings of the 34th International Conference on Machine Learning},
  pages = 	 {1321--1330},
  year = 	 {2017},
  editor = 	 {Precup, Doina and Teh, Yee Whye},
  volume = 	 {70},
  series = 	 {Proceedings of Machine Learning Research},
  month = 	 {06--11 Aug},
  publisher =    {PMLR},
  url = 	 {https://proceedings.mlr.press/v70/guo17a.html},
}

@inproceedings{geifman2017selective,
 author = {Geifman, Yonatan and El-Yaniv, Ran},
 booktitle = {Advances in Neural Information Processing Systems},
 editor = {I. Guyon and U. Von Luxburg and S. Bengio and H. Wallach and R. Fergus and S. Vishwanathan and R. Garnett},
 publisher = {Curran Associates, Inc.},
 title = {Selective Classification for Deep Neural Networks},
 url = {https://proceedings.neurips.cc/paper_files/paper/2017/file/4a8423d5e91fda00bb7e46540e2b0cf1-Paper.pdf},
 volume = {30},
 year = {2017}
}

@article{jackson2025omniscience,
  title = {{AA-Omniscience}: Evaluating Cross-Domain Knowledge Reliability in Large Language Models},
  author = {Jackson, Declan and Keating, William and Cameron, George and Hill-Smith, Micah},
  journal = {arXiv preprint arXiv:2511.13029},
  year = {2025},
  url = {https://arxiv.org/abs/2511.13029},
}

@inproceedings{recht2019imagenet,
  title = 	 {Do {I}mage{N}et Classifiers Generalize to {I}mage{N}et?},
  author =       {Recht, Benjamin and Roelofs, Rebecca and Schmidt, Ludwig and Shankar, Vaishaal},
  booktitle = 	 {Proceedings of the 36th International Conference on Machine Learning},
  pages = 	 {5389--5400},
  year = 	 {2019},
  editor = 	 {Chaudhuri, Kamalika and Salakhutdinov, Ruslan},
  volume = 	 {97},
  series = 	 {Proceedings of Machine Learning Research},
  month = 	 {09--15 Jun},
  publisher =    {PMLR},
  url = 	 {https://proceedings.mlr.press/v97/recht19a.html},
}

@misc{passos2026decisionindex,
  title = {{Decision Index 0.4.1}},
  author = {Apolin{\'a}rio Passos},
  year = {2026},
  howpublished = {Hugging Face Space, \url{https://huggingface.co/spaces/multimodalart/jev-decision-index}},
  note = {Edition 0.4.1. Leaderboard data generated 22 October 2026, 05:40 UTC; Space revision d9962113a1acc95c2d7522c2babfb989afbeb034. Accessed 22 October 2026},
}

@misc{aa2026decisionindex,
  title = {{Artificial Analysis Decision Index}},
  author = {{Artificial Analysis}},
  year = {2026},
  howpublished = {\url{https://artificialanalysis.ai/evaluations/artificial-analysis-decision-index}},
  note = {Version 1.0 (October 2026). Accessed 2026-10-21},
}

@article{cais2026hle,
  title = {A benchmark of expert-level academic questions to assess {AI} capabilities},
  author = {{Center for AI Safety} and {Scale AI} and {HLE Contributors Consortium}},
  journal = {Nature},
  volume = {649},
  number = {8099},
  pages = {1139--1146},
  year = {2026},
  doi = {10.1038/s41586-025-09962-4},
  url = {https://doi.org/10.1038/s41586-025-09962-4},
}

@inproceedings{ross2025when2call,
    title = "{W}hen2{C}all: When (not) to Call Tools",
    author = "Ross, Hayley  and
      Mahabaleshwarkar, Ameya Sunil  and
      Suhara, Yoshi",
    editor = "Chiruzzo, Luis  and
      Ritter, Alan  and
      Wang, Lu",
    booktitle = "Proceedings of the 2025 Conference of the Nations of the {A}mericas Chapter of the {A}ssociation for {C}omputational {L}inguistics: Human Language Technologies (Volume 1: Long Papers)",
    month = apr,
    year = "2025",
    address = "Albuquerque, New Mexico",
    publisher = "Association for Computational Linguistics",
    url = "https://aclanthology.org/2025.naacl-long.174/",
    doi = "10.18653/v1/2025.naacl-long.174",
    pages = "3391--3409",
    ISBN = "979-8-89176-189-6"
}

@inproceedings{han2024wildguard,
 author = {Han, Seungju and Rao, Kavel and Ettinger, Allyson and Jiang, Liwei and Lin, Bill Yuchen and Lambert, Nathan and Choi, Yejin and Dziri, Nouha},
 booktitle = {Advances in Neural Information Processing Systems},
 doi = {10.52202/079017-0261},
 editor = {A. Globerson and L. Mackey and D. Belgrave and A. Fan and U. Paquet and J. Tomczak and C. Zhang},
 pages = {8093--8131},
 publisher = {Curran Associates, Inc.},
 title = {{WildGuard}: Open One-stop Moderation Tools for Safety Risks, Jailbreaks, and Refusals of {LLMs}},
 url = {https://proceedings.neurips.cc/paper_files/paper/2024/file/0f69b4b96a46f284b726fbd70f74fb3b-Paper-Datasets_and_Benchmarks_Track.pdf},
 volume = {37},
 year = {2024}
}

@inproceedings{loshchilov2019decoupled,
  title = {Decoupled Weight Decay Regularization},
  author = {Loshchilov, Ilya and Hutter, Frank},
  booktitle = {International Conference on Learning Representations},
  year = {2019},
  url = {https://openreview.net/forum?id=Bkg6RiCqY7},
}

@misc{perplexity2026decider,
  title = {{pplx-decider-v1.1-27b}},
  author = {{Perplexity AI}},
  year = {2026},
  howpublished = {\url{https://huggingface.co/perplexity-ai/pplx-decider-v1.1-27b}},
  note = {Model card and released training and inference code (autojev), Apache-2.0. Accessed 2026-10-21},
}

@misc{baseten2026mammoth,
  title = {The top ranked decision models in the world: mammoth-1 and elephant-1},
  author = {Max Kirkby and Michael Psenka and Thomas Villeneuve and Aaryam Sharma and Charles O'Neill and Chloe Florit and Ardin Lo},
  year = {2026},
  howpublished = {\url{https://www.baseten.co/blog/the-best-decision-models-in-the-world-mammoth-1-and-elephant-1/}},
  note = {Baseten blog, 21 October 2026. Accessed 2026-10-22},
}

@article{yu2024distilling,
  title     = {Distilling System 2 into System 1},
  author    = {Yu, Ping and Xu, Jing and Weston, Jason and Kulikov, Ilia},
  journal   = {arXiv preprint arXiv:2407.06023},
  year      = {2024},
  url       = {https://arxiv.org/abs/2407.06023}
}

@article{deng2024explicit,
  title     = {From Explicit {CoT} to Implicit {CoT}: Learning to Internalize {CoT} Step by Step},
  author    = {Deng, Yuntian and Choi, Yejin and Shieber, Stuart},
  journal   = {arXiv preprint arXiv:2405.14838},
  year      = {2024},
  url       = {https://arxiv.org/abs/2405.14838}
}

@inproceedings{hsieh2023distilling,
  title     = {Distilling Step-by-Step! Outperforming Larger Language Models with Less Training Data and Smaller Model Sizes},
  author    = {Hsieh, Cheng-Yu and Li, Chun-Liang and Yeh, Chih-kuan and Nakhost, Hootan and Fujii, Yasuhisa and Ratner, Alex and Krishna, Ranjay and Lee, Chen-Yu and Pfister, Tomas},
  booktitle = {Findings of the Association for Computational Linguistics: {ACL} 2023},
  pages     = {8003--8017},
  year      = {2023},
  url       = {https://aclanthology.org/2023.findings-acl.507/}
}

@inproceedings{goyal2024think,
  title     = {Think before you speak: Training Language Models With Pause Tokens},
  author    = {Goyal, Sachin and Ji, Ziwei and Rawat, Ankit Singh and Menon, Aditya Krishna and Kumar, Sanjiv and Nagarajan, Vaishnavh},
  booktitle = {The Twelfth International Conference on Learning Representations},
  year      = {2024},
  url       = {https://openreview.net/forum?id=ph04CRkPdC}
}

@inproceedings{menon2021statistical,
  title     = {A Statistical Perspective on Distillation},
  author    = {Menon, Aditya K. and Rawat, Ankit Singh and Reddi, Sashank and Kim, Seungyeon and Kumar, Sanjiv},
  booktitle = {Proceedings of the 38th International Conference on Machine Learning},
  series    = {Proceedings of Machine Learning Research},
  volume    = {139},
  pages     = {7632--7642},
  year      = {2021},
  url       = {https://proceedings.mlr.press/v139/menon21a.html}
}

@inproceedings{robinson2023leveraging,
  title     = {Leveraging Large Language Models for Multiple Choice Question Answering},
  author    = {Robinson, Joshua and Rytting, Christopher Michael and Wingate, David},
  booktitle = {The Eleventh International Conference on Learning Representations},
  year      = {2023},
  url       = {https://openreview.net/forum?id=yKbprarjc5B}
}

@inproceedings{zheng2024large,
  title     = {Large Language Models Are Not Robust Multiple Choice Selectors},
  author    = {Zheng, Chujie and Zhou, Hao and Meng, Fandong and Zhou, Jie and Huang, Minlie},
  booktitle = {The Twelfth International Conference on Learning Representations},
  year      = {2024},
  url       = {https://openreview.net/forum?id=shr9PXz7T0}
}

@inproceedings{tian2023just,
  title     = {Just Ask for Calibration: Strategies for Eliciting Calibrated Confidence Scores from Language Models Fine-Tuned with Human Feedback},
  author    = {Tian, Katherine and Mitchell, Eric and Zhou, Allan and Sharma, Archit and Rafailov, Rafael and Yao, Huaxiu and Finn, Chelsea and Manning, Christopher},
  booktitle = {Proceedings of the 2023 Conference on Empirical Methods in Natural Language Processing},
  pages     = {5433--5442},
  year      = {2023},
  url       = {https://aclanthology.org/2023.emnlp-main.330/}
}

@inproceedings{xiong2024can,
  title     = {Can {LLM}s Express Their Uncertainty? An Empirical Evaluation of Confidence Elicitation in {LLM}s},
  author    = {Xiong, Miao and Hu, Zhiyuan and Lu, Xinyang and Li, Yifei and Fu, Jie and He, Junxian and Hooi, Bryan},
  booktitle = {The Twelfth International Conference on Learning Representations},
  year      = {2024},
  url       = {https://openreview.net/forum?id=gjeQKFxFpZ}
}

@inproceedings{kamath2020selective,
  title     = {Selective Question Answering under Domain Shift},
  author    = {Kamath, Amita and Jia, Robin and Liang, Percy},
  booktitle = {Proceedings of the 58th Annual Meeting of the Association for Computational Linguistics},
  pages     = {5684--5696},
  year      = {2020},
  url       = {https://aclanthology.org/2020.acl-main.503/}
}

@inproceedings{broder1997resemblance,
  title = {On the resemblance and containment of documents},
  author = {Broder, Andrei Z.},
  booktitle = {Proceedings. Compression and Complexity of {SEQUENCES} 1997},
  pages = {21--29},
  publisher = {IEEE Computer Society},
  address = {Salerno, Italy},
  year = {1997},
  doi = {10.1109/SEQUEN.1997.666900},
  url = {https://doi.org/10.1109/SEQUEN.1997.666900},
}

@inproceedings{hu2022lora,
  title = {{LoRA}: Low-Rank Adaptation of Large Language Models},
  author = {Hu, Edward J. and Shen, Yelong and Wallis, Phillip and Allen-Zhu, Zeyuan and Li, Yuanzhi and Wang, Shean and Wang, Lu and Chen, Weizhu},
  booktitle = {International Conference on Learning Representations},
  year = {2022},
  url = {https://openreview.net/forum?id=nZeVKeeFYf9},
}

@misc{qwen2026qwen38_27b,
  key = {Qwen3.8-27B},
  author = {{Qwen Team}},
  title = {{Qwen3.8-27B}},
  year = {2026},
  howpublished = {\url{https://huggingface.co/Qwen/Qwen3.8-27B}},
}

@inproceedings{zaratiana2024gliner,
  title     = {{GLiNER}: Generalist Model for Named Entity Recognition using Bidirectional Transformer},
  author    = {Zaratiana, Urchade and Tomeh, Nadi and Holat, Pierre and Charnois, Thierry},
  booktitle = {Proceedings of the 2024 Conference of the North American Chapter of the Association for Computational Linguistics: Human Language Technologies (Volume 1: Long Papers)},
  pages     = {5364--5376},
  year      = {2024}
}

@inproceedings{zaratiana2025gliner2,
  title     = {{GLiNER2}: Schema-Driven Multi-Task Learning for Structured Information Extraction},
  author    = {Zaratiana, Urchade and Pasternak, Gil and Boyd, Oliver and Hurn-Maloney, George and Lewis, Ash},
  booktitle = {Proceedings of the 2025 Conference on Empirical Methods in Natural Language Processing: System Demonstrations},
  pages     = {130--140},
  year      = {2025},
  doi       = {10.18653/v1/2025.emnlp-demos.10},
  note      = {arXiv:2507.18546}
}

@article{zaratiana2026gliguard,
  title   = {{GLiGuard}: Schema-Conditioned Classification for {LLM} Safeguard},
  author  = {Zaratiana, Urchade and Newhauser, Mary and Hurn-Maloney, George and Lewis, Ash},
  journal = {arXiv preprint arXiv:2605.07982},
  year    = {2026},
  url     = {https://arxiv.org/abs/2605.07982}
}

@misc{typesafe2026jev,
  author       = {{TypeSafe AI}},
  title        = {Introducing System One Models \& {Jev}},
  howpublished = {TypeSafe AI blog, \url{https://typesafe.ai/blog/introducing-system-one-models-and-jev}},
  year         = {2026},
  month        = sep,
  note         = {15 September 2026. Accessed 2026-10-31}
}

@article{dhankhar2026drex,
  title = {Towards Foundation Decision Models: {D}rex},
  author = {Dhankhar, Nischay and Amantay, Alisher and Datbayev, Zhanibek and Akhmetov, Darkhan and Shah, Yash and Ubaidullayev, Nurmukhamed and Saparkhan, Raman and Baha, Dos and Saparov, Abulhair},
  journal = {arXiv preprint arXiv:2611.07854},
  year = {2026},
}

@misc{aa2026decisionindexv2,
  title = {Introducing the {Artificial Analysis Decision Index} v2},
  author = {{Artificial Analysis}},
  year = {2026},
  howpublished = {\url{https://artificialanalysis.ai/articles/decision-index-v2}},
  note = {8 November 2026. Accessed 2026-11-15},
}

@misc{passos2026decisionindexnov,
  title = {{Decision Index 0.4.1}},
  author = {Apolin{\'a}rio Passos},
  year = {2026},
  howpublished = {Hugging Face Space, \url{https://huggingface.co/spaces/multimodalart/jev-decision-index}},
  note = {Edition 0.4.1. Leaderboard data generated 12 November 2026, 12:34 UTC; Space revision 44be00f395c5af148320e9781945fe6ef4351283. Accessed 15 November 2026},
}

@inproceedings{dehghani2019universal,
  title = {Universal Transformers},
  author = {Dehghani, Mostafa and Gouws, Stephan and Vinyals, Oriol and Uszkoreit, Jakob and Kaiser, {\L}ukasz},
  booktitle = {International Conference on Learning Representations},
  year = {2019},
}

@inproceedings{bai2019deep,
  title = {Deep Equilibrium Models},
  author = {Bai, Shaojie and Kolter, J. Zico and Koltun, Vladlen},
  booktitle = {Advances in Neural Information Processing Systems},
  year = {2019},
}

@article{geiping2025scaling,
  title = {Scaling up Test-Time Compute with Latent Reasoning: A Recurrent Depth Approach},
  author = {Geiping, Jonas and McLeish, Sean and Jain, Neel and Kirchenbauer, John and Singh, Siddharth and Bartoldson, Brian R. and Kailkhura, Bhavya and Bhatele, Abhinav and Goldstein, Tom},
  journal = {arXiv preprint arXiv:2502.05171},
  year = {2025},
}

@inproceedings{saunshi2025reasoning,
  title = {Reasoning with Latent Thoughts: On the Power of Looped Transformers},
  author = {Saunshi, Nikunj and Dikkala, Nishanth and Li, Zhiyuan and Kumar, Sanjiv and Reddi, Sashank J.},
  booktitle = {International Conference on Learning Representations},
  year = {2025},
}

@misc{zaratiana2026gliner2piimultilingualmodelpersonally,
      title={GLiNER2-PII: A Multilingual Model for Personally Identifiable Information Extraction}, 
      author={Urchade Zaratiana and Ash Lewis and George Hurn-Maloney},
      year={2026},
      eprint={2605.09973},
      archivePrefix={arXiv},
      primaryClass={cs.CL},
      url={https://arxiv.org/abs/2605.09973}, 
}

\appendix

\section{\new{From a language model to a decision model, step by step}}
\label{a:walkthrough}

\new{This appendix explains in plain terms how \dmtwo{} is built from its base language model. It assumes only a basic knowledge of transformers. Sections~\ref{sec:architecture} and~\ref{sec:training} give the formal version, and Table~\ref{t:walkthrough} summarizes the changes.}

\subsection{\new{The starting point}}

\new{\basemodel{} is an ordinary language model: it reads a sequence of tokens and predicts the next one. It has 64 layers and a hidden size of 5,120, and its last layer gives a score to each of about 248,000 vocabulary tokens. The layers are of two kinds.}
\begin{itemize}[leftmargin=*,itemsep=1pt,topsep=2pt]
    \item \new{\textbf{16 self-attention layers.} In each, every token looks back at the earlier tokens and combines what it finds.}
    \item \new{\textbf{48 Gated DeltaNet layers.} Each reads the sequence from left to right and carries a running summary, much like a recurrent network. Its cost grows linearly with the length of the input, which keeps long documents cheap.}
\end{itemize}
\new{Both kinds are \emph{causal}: a token can use only what comes before it. A model that writes text one token at a time needs this, because the later tokens do not exist yet.}

\subsection{\new{The goal}}

\new{A decision model answers a different kind of request. It receives a context, a question and a list of options, and returns one probability per option. For example:}
\begin{quote}
\small
\new{\textbf{Context:} Order 1182. The parcel arrived damaged; two photos attached.\\
\textbf{Question:} What should the agent do next?\\
\textbf{Options:} (A) refund \ (B) replace \ (C) escalate \ (D) ask for more information\\
\textbf{Wanted output:} A 0.64, \ B 0.21, \ C 0.11, \ D 0.04}
\end{quote}
\new{A language model could answer this by writing text, but that has three drawbacks. Writing is slow, because every generated token needs another pass through the network. The text has to be parsed. And the probabilities of the options are not given directly. Letting the model reason step by step before it answers improves accuracy, but multiplies the cost.}

\new{We want the accuracy of the reasoning model at the cost of a single pass. The six steps below get there.}

\subsection{\new{Step 1: write the request as one prompt}}

\new{The context, the question and the options are placed in a single prompt. Each option is given a one-token code (A, B, C, \ldots). The prompt ends with the prefix ``\texttt{Answer:~(}'', so that the next token a language model would write is the code of its answer.}

\subsection{\new{Step 2: read the answer from one position}}

\new{The network turns the prompt into one vector per token. We keep only the vector at the last token, $h$, which has 5,120 numbers. A small matrix $W$ with 255 rows, one for each possible option code, turns $h$ into 255 scores, $z=Wh$. We keep the scores of the options that were actually offered (four in the example) and apply a softmax. The result is one probability per option, and the probabilities sum to one.}

\new{This readout replaces the original output layer, which scored 248,000 tokens. $W$ does not start from random values: each row is copied from the original output layer's row for the same code token. So before any training, the readout scores option A exactly as the language model scored the token A.}

\new{There are 255 rows because we use 255 single-token codes: the 26 letters, then two-letter codes. A question can therefore have at most 255 options.}

\subsection{\new{Step 3: let every token see the whole prompt}}

\new{In a language model, the attention mask lets token $i$ attend only to tokens $1,\ldots,i$. For a decision, this restriction is unnecessary, because the whole prompt is known before the model starts. It is also harmful: the context is read before the question has been seen, and option A is read before options B, C and D.}

\new{We remove the mask in the 16 self-attention layers. Every token can then attend to every other token. The model can read the context knowing the question, and it can compare the options with each other. We leave the 48 Gated DeltaNet layers unchanged. They are recurrent by design, and leaving them alone keeps their linear cost.}

\new{Two consequences follow. First, the model can no longer write text, because writing depends on the mask. Second, the pretrained weights have never seen ``future'' tokens, so the model has to be trained before it can use them. That is the job of the next step.}

\subsection{\new{Step 4: train on probabilities, not only on answers}}

\new{Each training example consists of a prompt, the correct option, and a \emph{target distribution} over the options, supplied by one or more teacher models. The loss (Equation~\ref{eq:training_loss}) adds two terms: 0.3 times the cross-entropy with the correct option, and 0.7 times the KL divergence from the target distribution.}

\new{The second term is what separates this from ordinary classification training. A label says only that A is right. A target such as $(0.70,\,0.05,\,0.25,\,0.00)$ also says that C is a plausible alternative and that D is not. A student trained on such targets learns how sure to be, not only what to answer. This is where the model's calibration comes from. No temperature or other correction is fitted after training.}

\subsection{\new{Step 5: move the thinking into training}}

\new{One teacher is a version of the base model itself that reasons before it answers (Stage~1 in Section~\ref{sec:training}). For each training question we let this teacher write out its reasoning, and then record the probability it gives to each option. The student never sees the reasoning text. It sees only the distribution that the teacher reached after thinking, and learns to produce that distribution directly.}

\new{The cost of thinking is therefore paid once, during training, and not on every request. Thinking does not help on every question. If the teacher is already confident without reasoning, we use its direct distribution. Otherwise, we average its direct and post-thinking distributions (Equation~\ref{eq:reason_target}).}

\subsection{\new{Step 6: refine in rounds and average the weights}}

\new{After the conversion, five short training rounds follow. After every round, we average the new weights with the weights the round started from. This keeps what earlier rounds learned. A final stage retrains the model on a rebuilt data mixture, with targets pooled from several teachers (Section~\ref{sec:training}). The released model is the average of two such runs with different random seeds.}

\subsection{\new{What comes out}}

\new{The finished model takes a prompt and returns one probability per option, in a single pass that takes a median of 92.9\,ms. An application can take the most probable option. It can also act only when that probability is high enough, and pass the other cases to a person (Appendix~\ref{a:primer}).}

\subsection{\new{After GLiDE V1.2}}

\new{Three later steps build on the finished model (Section~\ref{s:later}). None changes the picture above: the model still returns one probability per option in a single pass.}
\begin{itemize}[leftmargin=*,itemsep=1pt,topsep=2pt]
    \item \new{\textbf{Average in two more models.} Two further models were trained on broader data. Their weights were averaged with those of \dmtwo{}.}
    \item \new{\textbf{Run some layers twice.} GLiDE~27B applies two blocks of middle layers twice at inference, moving half-way each time. The model computes more per question with no new weights and no generated tokens.}
    \item \new{\textbf{Teach smaller models.} Smaller models of the same design repeat the recipe. They are then trained to match the output distributions of the 27B model.}
\end{itemize}

\begin{table}[t]
\centering
\footnotesize
\setlength{\tabcolsep}{3pt}
\renewcommand{\arraystretch}{1.08}
\begin{tabularx}{\columnwidth}{
@{}p{0.24\columnwidth}
>{\raggedright\arraybackslash}X
>{\raggedright\arraybackslash}X@{}
}
\toprule
\textbf{Component} & \textbf{Base LM} & \textbf{\dmtwo{}} \\
\midrule
Input
& Text prefix
& Context, question, options \\

Full attention (16)
& Causal
& Bidirectional \\

Gated DeltaNet (48)
& Causal
& Causal \\

Readout
& $\sim$248K token logits
& 255 option logits \\

Prediction
& Next token
& Option probabilities \\

Forward passes
& One per token
& One per decision \\

Supervision
& Next-token prediction
& Gold and soft targets \\
\bottomrule
\end{tabularx}
\caption{\textbf{From a language model to \dmtwo{}.}
The full-attention mask and readout change, while
Gated DeltaNet layers retain causal recurrence.}
\label{t:walkthrough}
\end{table}

\section{Related Work}
\label{s:related}

\paragraph{Distillation and one-pass reasoning.}
Knowledge distillation transfers a teacher's soft distribution to a student~\citep{hinton2015distilling,kim2016sequence}, including information in the probability mass assigned to incorrect answers~\citep{menon2021statistical}. Prior work has compressed chain-of-thought reasoning through sampled rationales, final answers, removed reasoning steps, or latent tokens~\citep{hsieh2023distilling,yu2024distilling,deng2024explicit,goyal2024think}. In contrast, we distill the complete post-thinking answer distribution, preserving the teacher's residual uncertainty.

\paragraph{Single-pass decision models.}
Schema-conditioned encoders such as GLiNER, GLiNER2, and GLiGuard support efficient prediction over natural-language labels or schemas~\citep{zaratiana2024gliner,zaratiana2025gliner2,zaratiana2026gliner2piimultilingualmodelpersonally,zaratiana2026gliguard}. Work on non-causal decoders removes causal masks to produce strong representations~\citep{behnamghader2024llm2vec,lee2025nvembed}; in our hybrid decoder, we unmask only the full-attention layers. Because option-symbol scoring can introduce order bias~\citep{robinson2023leveraging,zheng2024large}, we use a dedicated readout and shuffle options during training. \new{Drex Max~\citep{dhankhar2026drex}, developed concurrently, uses the same backbone, also removes the causal mask, and also adds a 255-way readout. Its training differs: it adds consistency losses and a reinforcement-learning stage.}

\paragraph{Calibration and abstention.}
Neural networks and language models are often overconfident~\citep{guo2017calibration,tian2023just,xiong2024can}. Selective prediction therefore adds a reject mechanism~\citep{chow1970optimum,geifman2017selective,kamath2020selective}; we train explicit abstention as an ordinary option and expose probabilities for application-level deferral.

\section{Decision-model output semantics}
\label{a:primer}
Section~\ref{sec:architecture} defines how \dmtwo{} maps a context, question, and finite answer set to a categorical distribution. This appendix specifies how that distribution is interpreted by an application. It also makes explicit several operational properties that are separate from the network architecture: typed questions, cost-sensitive decisions, deferral, permutation equivariance, and the independence of batched questions.

\subsection{Typed questions}

For question $j$, let
\[
    p_j=(p_{j,1},\ldots,p_{j,K_j})
    \in\Delta^{K_j-1}
\]
be the distribution from Equation~\ref{eq:decision}. The same distribution supports three output types.

\paragraph{Choice.}
For an unordered set of named actions, the default decision and its confidence are
\[
    \hat{k}_j=\arg\max_{k\in[K_j]}p_{j,k},
    \qquad
    \hat{p}_j=p_{j,\hat{k}_j}.
\]
The option descriptions are part of the input, so the actions need not belong to a fixed global label vocabulary. The model can score at most 255 options because its readout contains 255 single-token option codes.

\paragraph{Yes/no.}
If the two options are ``yes'' and ``no'', the useful scalar is
\[
    s_j=p_j(\mathrm{yes}),
    \qquad
    p_j(\mathrm{no})=1-s_j.
\]
A threshold of $1/2$ is optimal only when false positives and false negatives have equal cost. If their costs are respectively $C_{\mathrm{FP}}$ and $C_{\mathrm{FN}}$, the minimum-expected-cost rule predicts yes when
\[
    s_j \geq
    \frac{C_{\mathrm{FP}}}
         {C_{\mathrm{FP}}+C_{\mathrm{FN}}}.
\]
Thus an application can change its operating point without retraining the model.

\paragraph{Ordered score.}
Suppose the options represent ordered numerical levels
$v_1<\cdots<v_{K_j}$. The modal and expected scores are
\[
    \hat{v}^{\,\mathrm{mode}}_j
    =v_{\arg\max_k p_{j,k}},
    \qquad
    \hat{v}^{\,\mathrm{mean}}_j
    =\sum_{k=1}^{K_j}p_{j,k}v_k.
\]
The full distribution also exposes ambiguity through
\[
    \operatorname{Var}(v_j)
    =\sum_{k=1}^{K_j}p_{j,k}
      \left(v_k-\hat{v}^{\,\mathrm{mean}}_j\right)^2.
\]
For example, $(0.05,0.10,0.20,0.45,0.20)$ over levels
$0,\ldots,4$ has mode $3$, mean $2.65$, and retains the information that adjacent levels remain plausible.

\subsection{Automation, deferral, and abstention}

The model distribution and the application's decision policy are distinct. A common selective policy~\citep{chow1970optimum,geifman2017selective} automates only sufficiently confident predictions:
\[
    \delta_\tau(u_j)=
    \begin{cases}
        \hat{k}_j, & \hat{p}_j\geq\tau,\\
        \mathrm{defer}, & \hat{p}_j<\tau.
    \end{cases}
\]
On an evaluation set of $N$ questions, this policy has coverage
\[
    \operatorname{Cov}(\tau)
    =\frac{1}{N}\sum_{j=1}^{N}
      \mathbf{1}[\hat{p}_j\geq\tau]
\]
and selective risk
\[
    \operatorname{Risk}(\tau)
    =
    \frac{
      \sum_j\mathbf{1}[\hat{p}_j\geq\tau]
      \mathbf{1}[\hat{k}_j\neq y_j]}
    {
      \sum_j\mathbf{1}[\hat{p}_j\geq\tau]
    }.
\]
Increasing $\tau$ usually lowers coverage and aims to lower risk; calibration determines whether confidence has a stable operational meaning on the deployment population. Deferral is not the same as explicit abstention. Deferral is imposed by the caller after inference and can use any threshold. Explicit abstention instead adds an ``I don't know'' option $\bot$ to the candidate set, so that
\[
    p_j\in\Delta^{K_j},
    \qquad
    \hat{k}_j\in\{1,\ldots,K_j,\bot\}.
\]
It is therefore a behavior learned from training targets, not a post-hoc confidence rule. An application may combine both mechanisms: accept $\bot$ when the model selects it, and defer any remaining answer below a chosen confidence threshold.

\subsection{Batched questions and independence}

A request may contain $M$ questions sharing a context $c$. Each is rendered as its own sequence,
\[
    x_j=\operatorname{Tok}(c,q_j,\mathcal{A}_j),
    \qquad j=1,\ldots,M,
\]
and the sequences are padded and evaluated in one batch:
\[
    (p_1,\ldots,p_M)
    =
    \operatorname{BatchEval}_\theta(x_1,\ldots,x_M).
\]
Batching shares a device call but does not couple the answer distributions: the API exposes $M$ marginals rather than a joint distribution. If an application interprets them independently, the induced joint distribution is
\[
    p(\hat{k}_1,\ldots,\hat{k}_M\mid\mathcal{R})
    =
    \prod_{j=1}^{M}p_j(\hat{k}_j\mid c,q_j,\mathcal{A}_j).
\]
Consequently, cross-question constraints---for example, requiring exactly one of several flags to be true---must be enforced by a downstream constrained decoder or represented as one joint choice question. The context is included once within each question sequence, but is repeated across the $M$ batched sequences.

\subsection{Option order and semantic equivariance}

Candidate order should not change the underlying decision. For a permutation $\pi$ of the $K_j$ options, the desired equivariance property is
\[
    p_\theta\!\left(
      \pi(k)\mid c,q_j,\pi(\mathcal{A}_j)
    \right)
    =
    p_\theta(k\mid c,q_j,\mathcal{A}_j).
\]
This property is not guaranteed by the architecture because each option is paired with an ordered code and occupies a different prompt position. Shuffling options during training, while applying the same permutation to the target distribution, encourages approximate equivariance. Keeping the full option descriptions in the prompt also makes the prediction depend on their semantics rather than on a fixed meaning for code A, B, and so on.

\begin{figure*}[t]
\centering
\begin{tikzpicture}
\node[anchor=south west, inner sep=0] (img) at (0,0) {\includegraphics[width=\textwidth]{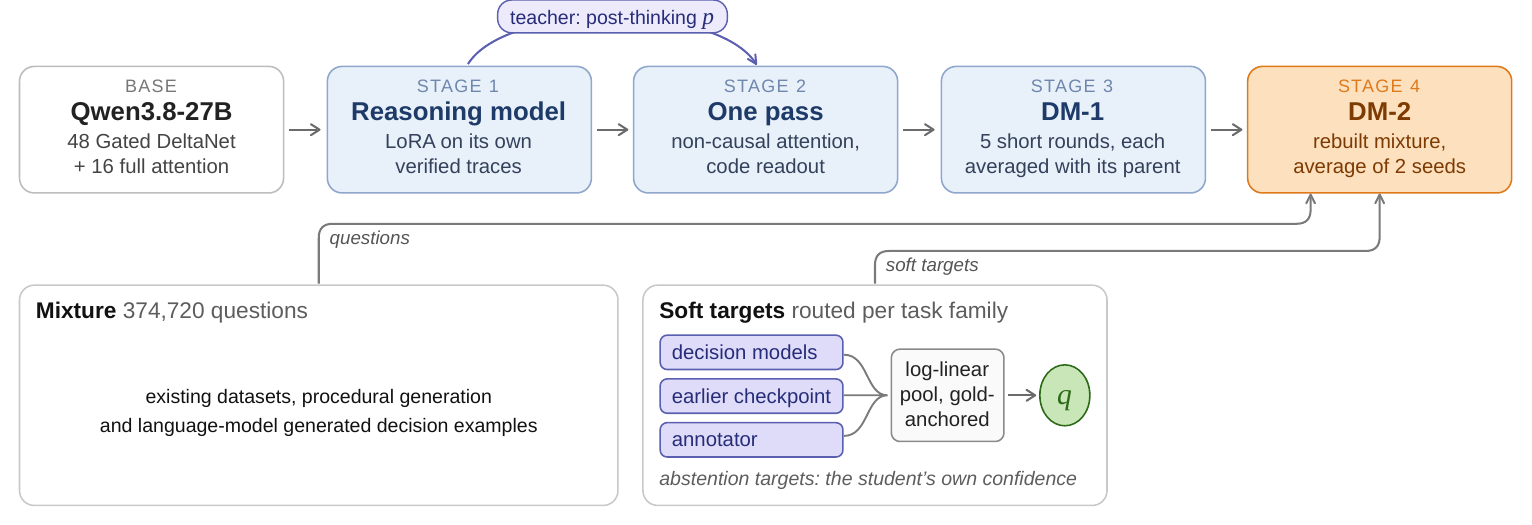}};
\begin{scope}[x={(img.south east)}, y={(img.north west)}]
\node[fill=pipeblue, inner sep=1.5pt, text=pipenavy, font=\sffamily\bfseries\fontsize{7.2}{8}\selectfont] at (0.6997,0.7883) {GLiDE V1.1};
\node[fill=pipeorange, inner sep=1.5pt, text=pipebrown, font=\sffamily\bfseries\fontsize{7.2}{8}\selectfont] at (0.8999,0.7896) {GLiDE V1.2};
\end{scope}
\end{tikzpicture}
\caption{\textbf{How \dmtwo{} was made.} Each stage continues from the weights of the one before; the Stage~1 reasoning model's post-thinking answer distribution is one of the soft targets for the single-pass conversion. Stage~4 trains \dmone{} on the rebuilt mixture with routed, gold-anchored soft targets and releases the average of two seeds.}
\label{f:pipe}
\end{figure*}

\section{The pipeline}
\label{a:pipe}
Figure~\ref{f:pipe} gives the end-to-end lineage from the pretrained base model to the released checkpoint, including staged post-training and the construction of supervision from routed teachers.

\subsection{Checkpoint lineage}

Let $\theta_0$ denote the \basemodel{} parameters. The four stages form a directed checkpoint lineage rather than four independent training runs:
\begin{equation}
\begin{aligned}
\theta_1
    &=\mathcal{T}_{\mathrm{reason}}
      (\theta_0;D_{\mathrm{reason}}),\\
\theta_2
    &=\mathcal{T}_{\mathrm{convert}}
      (\theta_1;D_{\mathrm{convert}},Q_{\mathrm{teacher}}),\\
\theta_3
    &=\mathcal{T}_{\mathrm{rounds}}^{(5)}(\theta_2)
      =\theta_{\mathrm{V1.1}},\\
\theta_{4,s}
    &=\mathcal{T}_{\mathrm{final}}
      (\theta_{\mathrm{V1.1}};D_{\mathrm{final}},Q_{\mathrm{routed}},s), s\in\{1,2\}.
\end{aligned}
\label{eq:pipeline_lineage}
\end{equation}
Here $\mathcal{T}_{\mathrm{reason}}$ produces the thinking-enabled teacher; $\mathcal{T}_{\mathrm{convert}}$ introduces non-causal full attention and the 255-way option readout while distilling post-thinking behavior; and $\mathcal{T}_{\mathrm{rounds}}^{(5)}$ denotes five short training rounds. In round $r$, the newly trained checkpoint $\widetilde{\theta}_{3,r}$ is averaged with the parent that initialized it,
\begin{equation}
    \theta_{3,r}
    =\tfrac{1}{2}
      \left(\theta_{3,r-1}+\widetilde{\theta}_{3,r}\right),
    \qquad r=1,\ldots,5,
\end{equation}
which limits forgetting while accumulating the updates. Both final seeds start from the same \dmone{} checkpoint; they differ only through training stochasticity. The released parameters are their uniform average,
\begin{equation}
    \theta_{\mathrm{V1.2}}
    =\tfrac{1}{2}
      \left(\theta_{4,1}+\theta_{4,2}\right).
\label{eq:final_average}
\end{equation}

\subsection{Supervision path}

The final stage does not train directly on raw labels alone. Its 374,720 questions first pass through format normalization and data-quality screening. Each retained row $i$ contains an input $x_i$, gold option $y_i$, task family $f_i$, and routed soft target $q_i$. For ordinary rows, $q_i$ is the cross-fitted, task-specific log-linear teacher pool of Equation~\ref{eq:teacher_pool}; incorrect teacher consensus is replaced by a gold-anchored target. Knowledge rows instead use the starting student's own correctness and confidence to decide whether the target should favor the answer or the explicit abstention option. At each optimization step, the candidate options, their codes, the gold index, and every coordinate of $q_i$ are permuted together. The student then minimizes Equation~\ref{eq:training_loss}. This ordering is important: teacher routing and gold anchoring define the target \emph{before} option shuffling, while the shared permutation preserves its semantics and prevents a code position from becoming a label shortcut.

\section{Development Evaluation}
\label{a:sel}
Internal development evaluation covered held-out task families and domains, input robustness, abstention, and regression checks. A composite development score summarized these measurements for checkpoint comparison. Official hidden evaluations remained externally administered and returned only aggregate final-model results.

\subsection{Statistical analysis}

Unless stated otherwise, bracketed intervals compare two models on the same items using paired percentile bootstrap resampling within each skill. The ECE interval is a parametric bootstrap from the board's confidence bins, and the regression interval in \S\ref{s:results} is a $t$ prediction interval. Counts written with K or M are rounded; all other counts are exact.

\subsection{Data-quality screening}
\label{a:data_quality}

Before final training, candidate rows were checked for exact and near duplicates across source and evaluation collections using question, passage, and option matching, shared eight-word sequences, and MinHash similarity~\citep{broder1997resemblance}. This removed 3,222 of 377,942 candidate rows. We did not replace them, leaving the 374,720-question final mixture reported in Table~\ref{t:data}.

\section{\new{Artificial Analysis Decision Index, all versions}}
\label{a:aa}
\new{Table~\ref{t:aa_full} lists every decision model on versions~1, 2 and~2.1 of the Artificial Analysis Decision Index. The scores are those published by Artificial Analysis. We retrieved version~1 on 5~November 2026, version~2 on 8~November (archived copy of the page) and version~2.1 on 15~November. Some developers have more than one model. The chart on the leaderboard shows these scores as whole numbers.}

\begin{table}[h]
\centering\footnotesize\setlength{\tabcolsep}{2.4pt}
\new{\begin{tabular}{@{}llrrr@{}}
\toprule
Model & Developer & v1 & v2 & v2.1 \\
\midrule
Drex Max v1.1 & Nace.AI & 63.78 & 73.39 & 72.62 \\
SubQ Snap 1.2 & Subquadratic & -- & -- & 72.58 \\
\textbf{GLiDE V1.2} & Fastino & 65.01 & 72.83 & 72.52 \\
Mammoth 27B & Baseten & -- & 72.45 & 71.36 \\
d1-ultra & Liquid AI & 65.25 & 70.45 & 70.58 \\
Decider v1.3 & Perplexity & -- & -- & 70.41 \\
Decider v1.2 & Perplexity & 56.17 & 70.48 & 69.77 \\
SubQ Snap 1.1 & Subquadratic & 65.00 & 69.17 & 67.92 \\
d1.1 & Liquid AI & -- & -- & 66.49 \\
Elephant 9B & Baseten & 64.69 & 67.55 & 66.39 \\
Jev 1.14 & TypeSafe & 57.07 & 66.96 & 66.23 \\
Gemini Decision 1 & Google & -- & -- & 65.98 \\
Decider Flash v1.3 & Perplexity & -- & -- & 64.07 \\
Microsoft-Decision-1 & Microsoft & 53.76 & 62.80 & 62.07 \\
Mercury Decide & Inception & 52.72 & 62.96 & 62.07 \\
Celeris-1.1 Decision & Celeris & 60.93 & 63.53 & 61.98 \\
Drex 2.0 & Nace.AI & 57.89 & 63.43 & 61.56 \\
d1 & Liquid AI & 50.29 & 61.59 & 60.78 \\
Clef & Cloudflare & 48.61 & 60.21 & 60.33 \\
GPT-6 Luna (Decisions) & OpenAI & 48.49 & 61.73 & 60.25 \\
Decider Flash v1.2 & Perplexity & 50.15 & 61.59 & 60.08 \\
Clef-Omni & Cloudflare & 44.26 & 59.27 & 57.75 \\
Drex Max & Nace.AI & 63.74 & -- & -- \\
Jev 1.13 & TypeSafe & 57.50 & -- & -- \\
GLiDE v1 (earlier) & Fastino & 53.91 & -- & -- \\
Decider v1.1 27B & Perplexity & 53.06 & -- & -- \\
Celeris-1 Decision & Celeris & 52.29 & -- & -- \\
Drex 1.5 & Nace.AI & 50.73 & -- & -- \\
Clef-Flash & Cloudflare & 45.81 & -- & -- \\
\bottomrule
\end{tabular}}
\caption{\new{\textbf{Artificial Analysis Decision Index, versions 1, 2 and 2.1:} all decision models, ordered by their latest score. Our model is in bold. A dash means the model was not listed on that version.}}
\label{t:aa_full}
\end{table}

\end{document}